\documentclass{article}
\usepackage{arxiv}
\usepackage{amsmath}
\usepackage[utf8]{inputenc}
\usepackage[T1]{fontenc}
\usepackage{url}            
\usepackage{booktabs}       
\usepackage{array}
\usepackage{amsfonts}       
\usepackage{amssymb}        
\usepackage{enumitem}       
\usepackage{nicefrac}       
\usepackage{microtype}      
\usepackage{lipsum}
\usepackage{graphicx}       
\usepackage{physics}
\usepackage{amsthm}
\usepackage[caption=false,font=footnotesize]{subfig}
\usepackage{hyperref}
\usepackage{color}
\usepackage{tikz}

\newtheorem{theorem}{Theorem}

\newtheorem{corollary}{Corollary}
\newtheorem{lemma}[theorem]{Lemma}

\newcommand{\plotplaceholder}[2]{%
    \IfFileExists{#1}{%
        \includegraphics[width=\linewidth]{#1}%
    }{%
        \fbox{%
            \begin{minipage}[c][3.5cm][c]{0.95\linewidth}
                \centering
                \small #2\\[0.5em]
                \footnotesize Figure to be generated
            \end{minipage}%
        }%
    }%
}

\newcommand{\circledblack}[1]{%
  \tikz[baseline=(char.base)]{%
    \node[shape=circle, fill=black, text=white, inner sep=1pt] (char) {#1};%
  }%
}

\makeatletter
\newcommand\notsotiny{\@setfontsize\notsotiny{6.3}{8.5}}
\makeatother

\usepackage{cite}

\renewcommand{\shorttitle}{Topology and Data Heterogeneity in Decentralized Learning}

\begin{document}

\raggedbottom

\title{How Network Topology and Data Heterogeneity Shape Decentralized Learning: \\A Spectral Analysis}

\author{Luigi~Palmieri, Lorenzo~Valerio, Chiara~Boldrini, Andrea~Passarella, Marco~Conti\\
        IIT institute of the National Research Council, Pisa
        \And
        János~Kertész\\
        Central European University, Vienna}
\date{}

\maketitle

\begin{abstract}
Network topology influences decentralized federated learning through information mixing and the relative contribution of individual nodes. While existing analyses primarily focus on spectral mixing properties, they often assume doubly stochastic aggregation, overlooking topology-dependent influence under irregular communication networks. We develop a spectral framework that separates two complementary mechanisms: the dominant eigenmode determines the consensus trajectory and node influence, while non-dominant eigenmodes govern the propagation and attenuation of disagreement. Under uniform neighborhood averaging, we show that the stationary influence is proportional to node degree plus one and derive a bound relating model disagreement to spectral mixing and gradient heterogeneity. We then investigate how these mechanisms interact with the data distribution across nodes. Experiments across regular and heterogeneous network topologies reveal nearly identical learning dynamics under IID data, while network size and connectivity affect how unevenly distributed information is learned under non-IID data. For non-IID settings, we characterize how the distribution of data across nodes shapes the relative contribution of local information to consensus dynamics. 
\end{abstract}

\keywords{Decentralized federated learning \and network topology \and spectral graph analysis \and data heterogeneity \and stationary distribution}

\section{Introduction}
\label{sec:introduction}

Federated learning (FL) enables multiple devices or organizations to collaboratively train machine learning models without sharing their local data. While traditional FL architectures rely on a central server to coordinate training and aggregate model updates~\cite{ref:mcmahan_fedavg}, decentralized federated learning (DFL) removes this dependency by allowing nodes to exchange model parameters directly with their neighbors~\cite{lian_can_2017,koloskova2020unified}. This approach is particularly relevant in environments where centralized coordination is impractical or undesirable, such as peer-to-peer networks, edge computing systems, and distributed sensing infrastructures. However, decentralization also introduces new challenges, as the learning process becomes dependent on the communication network connecting the participating nodes.

Understanding how network topology affects decentralized learning has therefore attracted considerable attention. Existing theoretical analyses have investigated how the spectral properties of the communication matrix influence information mixing, model disagreement, and convergence rates~\cite{lian_can_2017,nedic2017achieving,koloskova2020unified}. In particular, the spectral gap has traditionally been used to characterize how quickly differences between local models are attenuated, although more recent analyses have shown that the complete eigenvalue spectrum can provide a more accurate characterization of topology-dependent convergence~\cite{takezawa2026topology}. Several studies have also identified conditions under which the impact of topology becomes limited~\cite{pu_asymptotic_2020,neglia2020decentralized}, while others have shown that heterogeneous data distributions can substantially alter its influence on learning~\cite{lebars2023topology,dandi_data-heterogeneity-aware_2022}. Nevertheless, many of these analyses assume doubly stochastic mixing matrices, which assign equal influence to all nodes in the consensus process. This assumption simplifies the analysis but does not necessarily reflect decentralized averaging over irregular communication networks.

Indeed, consider a decentralized system in which each node averages the models received from its neighbors, including its own, assigning equal weight to each contribution. Although the communication graph may be undirected, the resulting mixing matrix is generally not doubly stochastic: nodes with different degrees normalize their incoming contributions differently. Consequently, the consensus component of the learning dynamics does not necessarily correspond to a uniform average of the local models and updates. Instead, individual nodes contribute according to a topology-dependent stationary distribution. Recent work has investigated the theoretical properties of row-stochastic aggregation and its implications for decentralized optimization~\cite{liang2025rowstochastic,liu2026rowstochastic}. However, the consequences of these non-uniform stationary weights for the contribution of heterogeneous information located at different nodes remain less explicitly characterized. This raises a question that convergence-oriented analyses do not fully address: \textit{beyond determining how quickly information propagates, how does network topology determine which information has greater influence on the learning process?}

The answer becomes particularly relevant when data are heterogeneously distributed. In practical decentralized systems, different nodes may observe different populations, environments, or classes, and some information may be available only at a small subset of nodes. Under these conditions, the influence of local information depends not only on its statistical properties but also on the position of the nodes holding it within the communication network. Previous empirical work has shown that information available at highly connected nodes can propagate more effectively than information located at peripheral nodes, leading to substantial differences in learning performance~\cite{palmieri2024topology}. At the same time, heterogeneous local gradients can continuously introduce differences between models, whose propagation and attenuation depend on the spectral properties of the network~\cite{lebars2023topology,yuan2023heterogeneity}. Understanding decentralized learning therefore requires considering both the influence assigned to individual nodes and the dynamics of disagreement among them.

In this paper, we investigate how network topology and data distribution jointly determine the influence of local information in decentralized learning. While existing spectral analyses primarily characterize convergence and information mixing, they provide less insight into how information held by different nodes contributes to the learning process. We address this question by combining the spectral decomposition of DecAvg dynamics, i.e., the fully decentralized version of FedAvg aggregation function\cite{sun_decentralized_2023}, with a class-wise characterization of local gradients. This allows us to distinguish the effects of network mixing from those of topology-dependent node influence, and to quantify how the placement of heterogeneous information shapes the consensus trajectory. In particular, our analysis provides a theoretical explanation for the hub amplification and peripheral dilution effects previously observed empirically, identifying the conditions under which network structure can favor or suppress localized information.

The main contributions of this work are as follows:
    \begin{itemize}
    \item \textbf{Topology-dependent information influence.} We characterize how stationary influence weights interact with heterogeneous local data distributions, deriving a class-contribution ratio that quantifies how node position and data placement jointly determine the relative contribution of localized information to the consensus trajectory.
    \item \textbf{Spectral analysis of learning dynamics.} Building on established spectral properties of reversible mixing matrices, we decompose DecAvg dynamics into consensus and disagreement components, characterizing how stationary node influence, spectral mixing, and gradient heterogeneity jointly shape learning dynamics.
    \item \textbf{Experimental evaluation.} Through experiments on regular and heterogeneous network topologies, we investigate the predicted effects of topology and data placement, showing limited topology dependence under IID data and differences in the learning of localized information under non-IID data.
\end{itemize}
Our findings highlight that the impact of network topology on decentralized learning cannot be characterized solely through connectivity or spectral mixing rates. The stationary distribution introduces an additional mechanism through which topology influences learning, particularly when data are not uniformly distributed across nodes. This perspective is especially relevant in decentralized deployments where communication patterns are constrained by the underlying infrastructure and cannot necessarily be optimized or redesigned.

The remainder of the paper is organized as follows. Section~\ref{sec:relwork} reviews related work. Section~\ref{sec:system_model} introduces the system model and DecAvg algorithm, while Section~\ref{sec:model_derivation} derives its matrix representation. Section~\ref{sec:eigen_problem} develops the spectral decomposition and characterizes consensus and disagreement dynamics. Section~\ref{sec:scenarios} investigates the interplay between topology and data distribution, and Section~\ref{sec:topologies} presents the experimental evaluation. Section~\ref{sec:conclusion} concludes the paper.

\section{Related work}
\label{sec:relwork}

\begin{table}[t]
\centering
\caption{Comparison of related approaches to network topology and data heterogeneity in decentralized learning.}
\label{tab:related_work_comparison}
\renewcommand{\arraystretch}{1.15}
\setlength{\tabcolsep}{3pt}
\scriptsize
\begin{tabular}{@{}>{\raggedright\arraybackslash}p{0.075\textwidth}>{\raggedright\arraybackslash}p{0.18\textwidth}>{\raggedright\arraybackslash}p{0.175\textwidth}>{\raggedright\arraybackslash}p{0.245\textwidth}>{\raggedright\arraybackslash}p{0.26\textwidth}@{}}
\hline
\textbf{Work} &
\textbf{Main focus} &
\textbf{Mixing assumptions} &
\textbf{Treatment of topology and data heterogeneity} &
\textbf{Node influence and data placement}\\
\hline

\cite{lian_can_2017} &
Convergence and disagreement in decentralized SGD &
Doubly stochastic mixing &
Relates spectral connectivity to information mixing, disagreement, and convergence &
Assumes uniform stationary influence; heterogeneous data placement not explicitly analyzed.\\

\cite{nedic2017achieving} &
Convergence and communication complexity &
Network-dependent spectral properties &
Characterizes how network connectivity affects convergence and communication requirements &
Focuses on network-level convergence rather than node-specific contributions.\\

\cite{zhu2022topology} &
Generalization and algorithmic stability &
Spectral connectivity in decentralized optimization &
Investigates how network structure affects stability and generalization &
Does not explicitly characterize heterogeneous data placement.\\

\cite{pu_asymptotic_2020} &
Asymptotic network independence &
Decentralized SGD under suitable learning conditions &
Identifies regimes in which the influence of topology diminishes &
Studies aggregate learning behavior rather than node-specific stationary influence.\\

\cite{nedic_network_2018} &
Network-dependent convergence analysis &
Network-dependent mixing &
Develops convergence results accounting for communication properties &
Does not explicitly investigate topology-dependent data placement.\\

\cite{neglia2020decentralized} &
Topology independence in decentralized learning &
Regular graphs and homogeneous data settings &
Relates gradient variability to sensitivity to network connectivity &
Explains when topology becomes less influential without analyzing localized information.\\

\cite{dandi_data-heterogeneity-aware_2022} &
Heterogeneity-aware decentralized optimization &
Adaptive mixing based on local data &
Adjusts aggregation weights to mitigate heterogeneity-induced effects &
Optimizes communication rather than characterizing the influence of fixed node positions.\\

\cite{koloskova2020unified} &
Unified convergence analysis of decentralized SGD &
Symmetric, doubly stochastic mixing &
Quantifies the effects of spectral connectivity, local steps, and gradient variability &
Assumes uniform stationary influence.\\

\cite{takezawa2026topology} &
Full-spectrum convergence analysis &
Symmetric, doubly stochastic mixing &
Accounts for the complete spectrum rather than only the spectral gap &
Focuses on convergence bounds rather than topology-dependent node influence.\\

\cite{yuan2023heterogeneity} &
Topology dependence and heterogeneity correction &
Doubly stochastic mixing &
Studies how removing heterogeneity effects changes topology-dependent convergence &
Does not explicitly characterize stationary influence or localized class contributions.\\

\cite{lebars2023topology} &
Neighborhood heterogeneity and topology learning &
Doubly stochastic mixing; topology adaptation &
Characterizes interactions between communication neighborhoods and heterogeneous data &
Studies neighborhood-level heterogeneity and topology design rather than stationary node influence.\\

\cite{huang2022sampletopology} &
Unified analysis with sample-induced topology &
Augmented communication and sampling graph &
Connects stochastic optimization, data heterogeneity, and network structure &
Focuses on convergence and heterogeneity correction rather than stationary influence.\\

\cite{kempe2003gossip} &
Average consensus through push-sum &
Non-doubly-stochastic communication &
Studies distributed information aggregation over directed networks &
Uses auxiliary weights to recover uniform averaging.\\

\cite{ref:nedic_olshevsky_consensus} &
Distributed optimization via subgradient-push &
Column-stochastic mixing over directed graphs &
Establishes convergence under directed and time-varying communication &
Compensates for non-uniform influence through auxiliary normalization.\\

\cite{assran2019stochastic} &
Stochastic Gradient Push &
Column-stochastic mixing over asymmetric networks &
Combines stochastic optimization with directed communication &
Corrects mixing imbalance rather than analyzing heterogeneous data placement.\\

\cite{liang2025rowstochastic} &
Optimization over row-stochastic networks &
Row-stochastic mixing &
Develops convergence guarantees and gradient-tracking methods &
Addresses convergence under non-uniform mixing rather than class-specific stationary influence.\\

\cite{liu2026rowstochastic} &
Weighted decentralized optimization &
Row-stochastic mixing; stationary-weighted analysis &
Derives convergence bounds and conditions favoring row-stochastic aggregation &
Studies weighted objectives and stationary influence, but not localized class contributions.\\

\cite{palmieri2024topology} &
Empirical topology and data-placement effects &
Neighborhood averaging over heterogeneous graphs &
Investigates how network structure and data placement affect learning performance &
Empirically identifies hub amplification and peripheral dilution without an explicit spectral characterization.\\

\cite{badiemodiri2025initialisation} &
Topology and heterogeneous model initialization &
Decentralized averaging with uncoordinated initialization &
Studies how network structure affects the evolution of independently initialized models &
Focuses on initialization effects rather than class-specific information placement.\\

\hline
\textbf{This work} &
\textbf{Topology-dependent consensus influence and disagreement} &
\textbf{Row-stochastic, reversible mixing without weight correction} &
\textbf{Separates consensus and disagreement; relates spectral mixing to gradient heterogeneity} &
\textbf{Quantifies how node position and local class distribution jointly determine contributions to the consensus trajectory.}\\
\hline

\end{tabular}
\end{table}

The relationship between network topology and decentralized learning has been investigated from several perspectives, including spectral convergence analysis, data heterogeneity, aggregation protocols, and information propagation over complex networks. We review these research directions below, focusing on their treatment of topology-dependent learning dynamics and node influence.

\subsection{Spectral Analysis of Decentralized Learning}

Early analyses of decentralized stochastic gradient descent (D-SGD) established the importance of spectral connectivity for convergence and model disagreement~\cite{lian_can_2017,nedic2017achieving,nedic_network_2018}. Subsequent work extended these results to generalization and stability~\cite{zhu2022topology} and identified conditions under which the influence of network topology diminishes asymptotically~\cite{pu_asymptotic_2020}. Koloskova et al.~\cite{koloskova2020unified} provide a unified convergence analysis of decentralized SGD variants, showing how spectral connectivity, communication frequency, and gradient variability jointly affect learning performance. More recently, Takezawa et al.~\cite{takezawa2026topology} demonstrate that convergence bounds based solely on the spectral gap can be overly pessimistic, deriving a refined characterization involving the complete eigenvalue spectrum. These works primarily investigate convergence properties under communication assumptions that yield uniform stationary node influence, rather than the effects of topology-dependent weighting on heterogeneous local information.

\subsection{Network Topology and Data Heterogeneity}

The interaction between topology and data distribution has received considerable attention. Neglia et al.~\cite{neglia2020decentralized} investigate conditions under which learning performance becomes largely independent of network connectivity, while Yuan et al.~\cite{yuan2023heterogeneity} analyze how removing heterogeneity-induced effects changes the topology dependence of decentralized SGD convergence. Other approaches explicitly incorporate data heterogeneity into communication. Dandi et al.~\cite{dandi_data-heterogeneity-aware_2022} propose adaptive mixing strategies that account for differences between local data distributions, while Le Bars et al.~\cite{lebars2023topology} introduce \emph{neighborhood heterogeneity}, a quantity that captures the interaction between communication topology and local objectives. They use this characterization to derive convergence bounds and guide topology selection under heterogeneous data. Huang et al.~\cite{huang2022sampletopology} provide a complementary framework based on sample-induced topology, jointly modeling communication and stochastic optimization. While these studies establish that topology and data heterogeneity cannot be considered independently, their primary focus is on convergence analysis or adapting communication to mitigate heterogeneity, rather than characterizing how fixed network structures weight information held by different nodes.

\subsection{Non-Doubly-Stochastic Aggregation}

Several approaches address decentralized learning with non-doubly-stochastic communication matrices. Push-sum protocols~\cite{kempe2003gossip} compensate for unequal aggregation weights through auxiliary normalization variables, enabling average consensus over asymmetric networks. This principle has been extended to distributed optimization through subgradient-push~\cite{ref:nedic_olshevsky_consensus} and to stochastic decentralized learning through Stochastic Gradient Push~\cite{assran2019stochastic}. More recent studies analyze row-stochastic communication directly. Liang et al.~\cite{liang2025rowstochastic} derive convergence guarantees and develop gradient-tracking methods for row-stochastic networks. Particularly relevant to our work, Liu et al.~\cite{liu2026rowstochastic} develop a stationary-weighted Hilbert-space analysis of row-stochastic aggregation, establishing convergence bounds and identifying conditions under which it can outperform doubly stochastic alternatives. Our analysis builds on these established properties of row-stochastic aggregation but addresses a different question: how do the stationary weights induced by a fixed communication topology interact with heterogeneous data placement? Rather than developing new convergence guarantees or correcting non-uniform influence, we characterize how topology-dependent weights affect the contribution of class-specific information to the consensus trajectory.

\subsection{Network Structure and Information Propagation}

Beyond convergence analysis, the structural position of nodes can influence how information propagates during decentralized learning. Palmieri et al.~\cite{palmieri2024topology} empirically investigate the interaction between network topology and data placement, showing that information held by highly connected nodes can spread more effectively, while information located at peripheral nodes may be diluted during aggregation. Their analysis considers several structural properties, including degree and betweenness centrality, clustering, and community structure.The present work extends this empirical investigation by providing an analytical characterization of the underlying mechanisms. In particular, we connect the stationary influence induced by network topology to class-specific gradient contributions, deriving an explicit relationship between node connectivity, local class proportions, and the relative contribution of localized information to the consensus trajectory. A related but distinct direction is investigated by Badie-Modiri et al.~\cite{badiemodiri2025initialisation}, who study the effects of network topology on decentralized learning with heterogeneous model initialization. In contrast, we assume homogeneous initialization and focus on the interaction between network topology and heterogeneous data distributions.

Table~\ref{tab:related_work_comparison} summarizes the approaches discussed above. Our work builds on established spectral and stationary-weighted analyses of decentralized learning, but shifts the focus from convergence guarantees to the influence of heterogeneous information placement. By combining the stationary distribution of the mixing matrix with a class-wise decomposition of local gradients, we characterize how network structure and local data distributions jointly determine the contribution of different classes to the consensus trajectory. The analysis also connects these contributions to disagreement dynamics, providing a unified interpretation of the topology-dependent learning effects previously observed empirically in~\cite{palmieri2024topology}.

\section{System Model and Preliminaries}
\label{sec:system_model}

Clients are connected through a fixed, connected, undirected graph $\mathcal{G}(\mathcal{V},\mathcal{E})$, where $\mathcal{V}$ is the set of nodes and $\mathcal{E}$ is the set of communication links. Connectivity ensures that information can eventually propagate across the whole network, while undirectedness yields a symmetric communication pattern (though the induced mixing matrix $K$ will still be generally non-symmetric due to degree normalization).
Let $N = |\mathcal{V}|$ and $E = |\mathcal{E}|$. For each node $i\in\mathcal{V}$, we define its \emph{closed neighborhood} as $\mathcal{N}(i)\triangleq\{i\}\cup\{j:(i,j)\in\mathcal{E}\}.$
We denote by $A\in\mathbb{R}^{N\times N}$ the adjacency matrix of $\mathcal{G}$, with $A_{ij}=1$ if $(i,j)\in\mathcal{E}$ and $A_{ij}=0$ otherwise, and by $d_i=\sum_j A_{ij}$ the degree of node $i$.
Connectivity ensures that information can eventually propagate across the entire network.
Each node $i$ holds a local dataset $\mathcal{D}_i$ and maintains a local model. Time is divided into communication rounds indexed by $t=0,1,2,\dots$. Within each round, nodes may perform multiple local optimization steps.

We study an averaging-based decentralized learning scheme, which we refer to as \emph{DecAvg}.
Each node $i$ maintains a model parameter vector $w^{(i)}_{t;\beta}$, where $t$ indexes the communication round and $\beta\in{0,\dots,\tau}$ indexes the local update step within that round.
Decentralized averaging may use non-uniform aggregation weights, for example proportional to local dataset sizes, as in FedAvg~\cite{ref:mcmahan_fedavg}.
We therefore consider a generic weighted formulation that allows both non-uniform aggregation weights and non-uniform self-influence.
This general formulation underlies the analysis that follows.

At each communication round $t$, all nodes synchronously perform the following steps:
\begin{enumerate}[leftmargin=*]
    \item \textbf{Neighborhood averaging.}
    Each node $i$ aggregates the models received from its closed neighborhood:
    \begin{equation}
        w^{(i)}_{t;0} \leftarrow \sum_{j\in\mathcal{N}(i)}
        \frac{ \alpha_{ij}}{\sum_{k\in\mathcal{N}(i)}\alpha_{ik}}w^{(j)}_{t-1;\tau}
        \label{eq:model_update_basic}
    \end{equation}
    where $\alpha_{ij}$ represents the weight of the model coming from node $j$ to node $i$.

    \item \textbf{Local training.}
    Starting from $w^{(i)}_{t;0}$, node $i$ performs $\tau$ local optimization steps on
    $\mathcal{D}_i$, producing the updated model $w^{(i)}_{t;\tau}$.
\end{enumerate}

Each node performs $\tau$ local gradient steps per communication round with learning rate $\eta$. We allow stochastic gradients and optional momentum with parameter $\mu \in [0,1)$; when momentum is used, its state is reset to zero at the beginning of each round. Sections \ref{sec:model_derivation}-\ref{sec:eigen_problem} first derive the dynamics for this general setting. The subsequent interpretation of consensus, disagreement, and data placement focuses on one local step without momentum ($\tau = 1$ and $\mu = 0$), matching the experimental configuration.

\subsection{Stacked representation and notation}

Let $d$ denote the dimension of the model parameter vector, so that each local model satisfies
$w^{(i)}_{t;\beta}\in\mathbb{R}^{d}$. We collect all node models into the stacked (block) vector
\begin{equation}
\mathbf{w}_{t;\beta} \triangleq
\begin{bmatrix}
w^{(1)}_{t;\beta}\\
\vdots\\
w^{(N)}_{t;\beta}
\end{bmatrix}
\in \mathbb{R}^{Nd}.
\label{eq:stacked_state}
\end{equation}
The vector $\mathbf{w}_{t;\beta}$ therefore represents the complete network state in $\mathbb{R}^{Nd}$. Communication mixes models across nodes but acts identically on each of their $d$ parameter coordinates. We represent this operation by the lifted mixing matrix
\begin{equation}
    \widetilde{K} \triangleq K\otimes I_d\in\mathbb{R}^{Nd\times Nd},
    \label{eq:K_tilde}
\end{equation}
where $\otimes$ is the Kronecker product, $K\in\mathbb{R}^{N\times N}$ describes how nodes combine their neighbors' models (we define $K$ in the next section) and $I_d$ is the $d\times d$ identity matrix. Thus, each scalar entry $K_{ij}$ is replaced by the block $K_{ij}I_d$, so the same node-level mixing rule is applied independently to every model coordinate.

The eigenvalues of $\widetilde{K}=K\otimes I_d$ are precisely those of $K$, each repeated $d$ times. Note that, since $\widetilde{K}=K\otimes I_d$, its eigenvectors take the form $\phi\otimes v$ for any eigenvector $\phi$ of $K$ and any $v\in\mathbb{R}^{d}$; the model-space factor $v$ is therefore a spectator to the mixing dynamics, and in Section~\ref{sec:eigen_problem} we can safely suppress it when performing the eigendecomposition. Consequently, the spectral behavior of communication in the full space $\mathbb{R}^{Nd}$ is completely determined by the spectrum of $K$ in $\mathbb{R}^{N}$. We may therefore analyze the smaller matrix $K$ without losing any information about the communication dynamics.

\section{Matrix Representation of DecAvg}
\label{sec:model_derivation}
We now express the node-wise averaging rule in matrix form. This representation separates network mixing from local updates, making the role of communication topology in the learning dynamics explicit. All communication effects are captured by a single mixing matrix $K\in\mathbb{R}^{N\times N}$. Powers of $K$ describe how information propagates through repeated mixing, allowing mixing speed, consensus, and long-term node influence to be studied through the spectrum of $K$.

To accommodate non-uniform aggregation weights, let
$A_{\alpha}\in\mathbb{R}^{N\times N}$ denote the weighted adjacency matrix, with
$(A_{\alpha})_{ij}=\alpha_{ij}$ for $i\neq j$ when $(i,j)\in\mathcal{E}$ and zero
otherwise. We assume $\alpha_{ij}=\alpha_{ji}>0$ on each edge, so
$A_{\alpha}$ is symmetric and has the same support as the adjacency matrix $A$.
The uniform case is recovered by setting $A_{\alpha}=A$. Let $\Gamma\triangleq\operatorname{diag}(\alpha_{11},\ldots,\alpha_{NN})$ contain the self-loop weights ($\alpha_{ii} > 0$), and define
\[
    A'\triangleq A_{\alpha}+\Gamma,
    \qquad
    D'\triangleq\operatorname{diag}(A'\mathbf{1}).
\]
Thus, the $i$th diagonal entry of $D'$ is the total weight used by node $i$,
$d'_i=\sum_{j\in\mathcal{N}(i)}\alpha_{ij}$ with $\mathcal{N}(i)$ defined as the neighborhood of node $i$ plus itself. The mixing matrix is the row-normalized
weighted adjacency matrix
\begin{equation}
    K \triangleq (D')^{-1}A'.
    \label{eq:interaction_matrix}
\end{equation}
Equivalently,
\[
    K_{ij}=
    \begin{cases}
        \displaystyle\frac{\alpha_{ij}}{d'_i}, & j\in\mathcal{N}(i),\\[4pt]
        0, & \text{otherwise}.
    \end{cases}
\]
Each row of $K$ therefore contains exactly the normalized weights used by one node
in the averaging step. Although $A'$ is symmetric, the row-dependent normalization
generally makes $K$ non-symmetric. Using the lifted matrix
$\widetilde{K}=K\otimes I_d$ introduced above, Equation~\eqref{eq:model_update_basic}
becomes
\begin{equation}
    \mathbf{w}_{t;0} = \widetilde{K}\mathbf{w}_{t-1;\tau}.
    \label{eq:stacked_simple}
\end{equation}
Combining this averaging step with the local updates and unrolling the recursion over
$t$ communication rounds gives the following expression; the proof is provided in
Appendix~\ref{sec:proofs}.
\begin{theorem}[Matrix Form of DecAvg Dynamics]
\label{thm:matrix_general_formula}
For any $t\ge 1$, the stacked model vector after the averaging step satisfies
\begin{equation}
    \mathbf{w}_{t;0} = \widetilde{K}^{t}\mathbf{w}_{0;0}
    -\eta\sum_{l=0}^{t-1}\widetilde{K}^{t-l}
    \sum_{\beta=0}^{\tau-1}a_{\tau,\beta}\nabla F(\mathbf{w}_{l;\beta}),
    \label{eq:decavg_dynamics}
\end{equation}
Here, $\nabla F(\mathbf{w}_{l;\beta})\in\mathbb{R}^{Nd}$ stacks the gradients of all nodes at communication round $l$ after $\beta$ local steps, and
\begin{equation}
    a_{\tau,\beta}\triangleq\frac{1-\mu^{\tau-\beta}}{1-\mu}.
    \label{eq:momentum_coefficient}
\end{equation}
accounts for the accumulation of gradient contributions over $\tau$ local steps under momentum parameter $\mu\in[0,1)$. For $\mu=0$, we recover standard SGD, with $a_{\tau,\beta}=1$ for all $\beta$.
\end{theorem}

The following lemma summarizes standard properties of reversible random-walk matrices~\cite{ref:chung_spectral,ref:boyd_fastmixing}, specialized to the mixing matrix $K$.

\begin{lemma}[Properties of $K$]
\label{lemma:K_properties}
The matrix $K$ defined in \eqref{eq:interaction_matrix} satisfies:
\begin{enumerate}
    \item $K$ is row-stochastic and primitive, and therefore admits a unique stationary distribution to which its powers converge;
    \item $K=I_N-(D')^{-1}L_{\alpha}$, where
    $L_{\alpha}\triangleq D'-A'$ is the weighted graph Laplacian;
    \item under the assumed undirected graph and symmetric edge weights
    ($\alpha_{ij}=\alpha_{ji}$), $K$ is diagonalizable with real eigenvalues.
\end{enumerate}
\end{lemma}

Property (1) permits a Markov-chain interpretation of the communication process, with $K$ as the transition matrix. Because $K$ is obtained by row-normalizing the symmetric matrix $A'$, it satisfies the detailed-balance condition $d'_i K_{ij} = A'_{ij} = d'_j K_{ji},$ and its stationary distribution is therefore $\pi_i = d'_i \big/ \sum_{j=1}^{N} d'_j$. These stationary probabilities determine the nodes' long-term influence. In general, $K$ is not doubly stochastic. It is doubly stochastic precisely when all weighted degrees $d'_i$ are equal, as occurs for a regular graph with uniform edge and self-loop weights.
Property (2) connects the dynamics to the normalized weighted Laplacian. In the
uniform case, $L_{\alpha}$ reduces to the usual combinatorial Laplacian $L=D-A$ because
the self-loop terms cancel in $D'-A'$. Finally, Property (3) follows because $K$ is
similar to the symmetric matrix $(D')^{-1/2}A'(D')^{-1/2}$. Hence, the dynamics can be
decomposed into real eigenmodes, as developed in Section~\ref{sec:eigen_problem}.

\section{Spectral Decomposition of Decentralized Learning Dynamics}
\label{sec:eigen_problem}

Section~\ref{sec:model_derivation} shows that the communication topology enters the learning dynamics through the interaction matrix $K$.
Under our assumptions of an undirected graph and symmetric aggregation weights, $K$ is diagonalizable with real eigenvalues (Lemma~\ref{lemma:K_properties}). We can therefore decompose the dynamics into eigenmodes and study convergence, mixing, and long-term influence through the spectral properties of $K$.

\subsection{A primer on Dirac bra-ket notation}
\label{sec:braket}

We adopt Dirac bra-ket notation to express vector expansions and spectral decompositions compactly. Since all vectors in our analysis are real, a column vector $v$ is denoted by $\ket{v}$, while its transpose is written as $\bra{v}$. Accordingly, $\braket{u}{v}$ denotes the inner product between two vectors, and $\ketbra{u}{v}$ denotes their outer product.

For a diagonalizable matrix $M \in \mathbb{R}^{n\times n}$, let $\{\omega_j\}_{j=1}^{n}$ denote its eigenvalues, with corresponding right eigenvectors $\{\ket{\phi_j}\}$ and left eigenvectors $\{\bra{\psi_j}\}$, normalized such that $\braket{\psi_i}{\phi_j}=\delta_{ij}$ (where $\delta_{ij}$ is the Kronecker delta, which equals 1 when $i=j$ and 0 otherwise). The associated spectral projectors are defined as $P_j \triangleq \ket{\phi_j}\bra{\psi_j}$. These projectors satisfy $P_iP_j=\delta_{ij}P_j$ (i.e., $P_i P_i = P_i$ and $P_iP_j = 0$) and $\sum_j P_j=I$, allowing us to express $M$ as
\begin{equation}
    M=\sum_{j=1}^{n}\omega_j P_j.
    \label{eq:spectral_decomposition}
\end{equation}
Unlike orthogonal projectors, the projectors $P_j$ are generally oblique when $M$ is non-symmetric, since its left and right eigenvectors need not coincide. This distinction is relevant to our analysis because the interaction matrix $K$ is generally non-symmetric, even when the underlying communication graph is undirected.

\subsection{Stacked representation in bra-ket notation}
\label{sec:stacked_braket}

We now express the stacked model representation introduced in Section~3.4 using bra-ket notation. Recall that the node space is $\mathbb{R}^{N}$, while each local model belongs to $\mathbb{R}^{d}$. The complete system state therefore belongs to the tensor-product space $\mathbb{R}^{N}\otimes\mathbb{R}^{d}$.

Let $\ket{i}$ denote the standard basis vector associated with node $i$ (with entry $1$ in position $i$ and $0$ elsewhere). The stacked model parameters (Eq.\eqref{eq:stacked_simple}) can be written as
\begin{equation}
    \ket{w_{t;\beta}} = \sum_{i=1}^{N} \ket{i}\otimes\ket{w_{t;\beta}^{(i)}}.
    \label{eq:stacked_braket}
\end{equation}
Similarly, we can rewrite Eq.\eqref{eq:K_tilde} as
\begin{equation}
     \widetilde{K} = K\otimes I_d = \sum_{i,j=1}^{N} K_{ij}\ket{i}\bra{j}\otimes I_d.
    \label{eq:mixing_operator_braket}
\end{equation}
where, in each term, $\ket{i}\bra{j}$ transfers the contribution of node $j$ to node $i$, $K_{ij}$ weights it, and $I_d$ leaves its model coordinates unchanged.

Equation~\eqref{eq:mixing_operator_braket} highlights that the mixing matrix $\widetilde{K}$ combines model parameters across neighboring nodes according to the weights in $K$, while applying the same mixing rule independently to each model coordinate (as identity operator $I_d$ ensures). Consequently, the spectral properties of the mixing process are entirely determined by the smaller matrix $K\in\mathbb{R}^{N\times N}$, whose spectral decomposition can be extended to the full model space through the Kronecker product. This allows us to study information propagation through the eigenmodes of $K$, as we do in the next subsection.

\subsection{Eigenmode decomposition of DecAvg}
\label{sec:eigenmode_decomposition}

We now exploit the spectral properties of the interaction matrix $K$ to obtain a mode-by-mode representation of the learning dynamics.

By Lemma~\ref{lemma:K_properties}, $K$ is diagonalizable with real eigenvalues under our assumptions of an undirected communication graph and symmetric aggregation weights. Let $\{\omega_j\}_{j=1}^{N}$ denote its eigenvalues, with corresponding right and left eigenvectors $\{\ket{\phi_j}\}$ and $\{\bra{\psi_j}\}$, respectively, normalized such that $\braket{\psi_i}{\phi_j}=\delta_{ij}$. Using the spectral projectors $P_j=\ket{\phi_j}\bra{\psi_j}$ introduced in Section~\ref{sec:braket}, we can write
\begin{equation}
K=\sum_{j=1}^{N}\omega_j P_j, \qquad K^t=\sum_{j=1}^{N}\omega_j^t P_j.
\label{eq:K_spectral}
\end{equation}

Recall that $K\in\mathbb{R}^{N\times N}$ describes communication between nodes, while the lifted matrix $\widetilde{K}=K\otimes I_d\in\mathbb{R}^{Nd\times Nd}$ acts on the stacked model parameters. Although they operate on different spaces, their spectral decompositions are directly related: $\widetilde{K}$ has the same eigenvalues as $K$, each repeated $d$ times. In particular, the spectral projectors of $K$ can be lifted to the full model space by defining
\begin{equation}
\widetilde{P}_j \triangleq P_j\otimes I_d.
\label{eq:lifted_projectors}
\end{equation}
The operators $P_j$ and $\widetilde{P}_j$ thus represent the same spectral component of the communication dynamics, but act on different spaces: $P_j$ projects vectors in $\mathbb{R}^{N}$, whereas $\widetilde{P}_j$ projects stacked model vectors in $\mathbb{R}^{Nd}$. Consequently, the powers of the lifted communication matrix can be expressed as
\begin{equation}
\widetilde{K}^{\,t}=\sum_{j=1}^{N}\omega_j^t\widetilde{P}_j.
\label{eq:Ktilde_spectral}
\end{equation}
Substituting this expression into the matrix formulation of DecAvg derived in Theorem~1 yields the following result.

\begin{theorem}[Eigenmode decomposition of DecAvg]
\label{thm:eigenmode_decavg}
Under the assumptions of Lemma~\ref{lemma:K_properties}, the stacked model parameters after the averaging step at communication round $t$ satisfy
\begin{equation}
\begin{aligned}
\ket{w_{t;0}} ={}& \sum_{j=1}^{N}\omega_j^t\widetilde{P}_j\ket{w_{0;0}}\\
&-\eta\sum_{l=0}^{t-1}\sum_{j=1}^{N}\omega_j^{t-l}\widetilde{P}_j\sum_{\beta=0}^{\tau-1}a_{\tau,\beta}\ket{\nabla F(w_{l;\beta})}.
\end{aligned}
\label{eq:eigenmode_decavg}
\end{equation}
where $a_{\tau,\beta}$ is the momentum coefficient defined in Equation~\eqref{eq:momentum_coefficient}.
\end{theorem}

\begin{proof}
Since $K$ is diagonalizable, its powers admit the spectral decomposition in Equation~\eqref{eq:K_spectral}. By the properties of the Kronecker product, the same decomposition extends to $\widetilde{K}$ as shown in Equation~\eqref{eq:Ktilde_spectral}. Substituting this expression into the matrix formulation of DecAvg in Theorem~1 and rearranging the summations gives Equation~\eqref{eq:eigenmode_decavg}.
\end{proof}

Equation~\eqref{eq:eigenmode_decavg} expresses DecAvg dynamics as a superposition of spectral components, making explicit how network topology and local learning jointly determine the evolution of model parameters. Three key observations follow.

\circledblack{1} \textbf{Eigenmodes and information propagation.} Each term in the sums over $j$ corresponds to an eigenmode of the mixing matrix $K$, representing a distinct pattern of information propagation across the network. The eigenvalue $\omega_j$ determines how contributions are attenuated over successive communication rounds, through the factors $\omega_j^t$ and $\omega_j^{t-l}$. The projector $\widetilde{P}_j$, instead, determines how the initial models and local gradients contribute to that mode. Since these projectors depend on both the left and right eigenvectors of $K$, networks with identical eigenvalues may still exhibit different learning dynamics.
\circledblack{2} \textbf{Initial conditions and local learning.} The two terms on the right-hand side of Equation~\eqref{eq:eigenmode_decavg} distinguish the propagation of initial model parameters from the accumulation of local learning updates. The first term describes how the initial models evolve under repeated averaging, independently of subsequent learning. The second term accounts for the gradients computed during local training, weighted by the coefficients $a_{\tau,\beta}$ and propagated through the network according to the eigenmodes of $K$. Thus, while the first term depends only on the initial conditions and network topology, the second also reflects the local data distributions and their interaction with the network structure.
\circledblack{3} \textbf{Consensus and deviations from consensus.} The eigenmode associated with $\omega_1=1$ determines the \textit{consensus component} of the learning trajectory.  Since its contributions are not attenuated by successive averaging, it captures the cumulative influence of local updates on the shared trajectory. In contrast, the remaining eigenmodes, associated with $|\omega_j|<1$, describe \textit{deviations from consensus}. Their contributions from previous rounds decay over time, but heterogeneous local gradients may continuously excite these modes, generating persistent disagreement. The decomposition therefore separates two complementary effects of network topology: its influence on the consensus trajectory and its role in propagating and attenuating differences between local models.

\subsection{Consensus and Disagreement Dynamics}
\label{sec:consensus_disagreement}

We now characterize the consensus and disagreement components identified in Theorem~\ref{thm:eigenmode_decavg}, starting from the spectral properties of the dominant eigenmode. We first recall the spectral properties of the dominant eigenmode, which follow from standard results on irreducible stochastic matrices and reversible Markov chains~\cite{ref:chung_spectral}.

\begin{theorem}[Spectrum and dominant eigenvector of $K$]
\label{thm:dominant_eigenmode}
Under the assumptions of Lemma~\ref{lemma:K_properties}, all eigenvalues of $K$ are real and satisfy $\omega_j\in(-1,1]$. The eigenvalue $\omega_1=1$ is simple, and its corresponding right eigenvector can be chosen as $\ket{\phi_1}=\ket{1}$, where $\ket{1}$ denotes the all-ones vector in $\mathbb{R}^N$.
\end{theorem}

\begin{proof}
Since $K$ is row-stochastic, $K\ket{1}=\ket{1}$, so $\ket{1}$ is a right eigenvector associated with eigenvalue $1$. Under the assumptions of Lemma~\ref{lemma:K_properties}, $K$ is irreducible and has strictly positive diagonal entries. It is therefore primitive, implying that $\omega_1=1$ is simple and $|\omega_j|<1$ for all $j\geq 2$. The eigenvalues are real because $K$ is similar to a symmetric matrix.
\end{proof}

We normalize the corresponding left eigenvector such that $\braket{\psi_1}{1}=1$. The dominant spectral projector is therefore $P_1=\ket{1}\bra{\psi_1}$, and its lifted counterpart is $\widetilde{P}_1=P_1\otimes I_d$. Using Theorem~\ref{thm:eigenmode_decavg}, the learning dynamics can be decomposed as
\begin{equation}
\ket{w_{t;0}}=\ket{w_{t;0}}_C+\ket{w_{t;0}}_D,
\label{eq:consensus_transient_decomposition}
\end{equation}
where
\begin{equation}
\ket{w_{t;0}}_C=\widetilde{P}_1\left(\ket{w_{0;0}}-\eta\sum_{l=0}^{t-1}\sum_{\beta=0}^{\tau-1}a_{\tau,\beta}\ket{\nabla F(w_{l;\beta})}\right)
\label{eq:consensus_component_general}
\end{equation}
and
{\small
\begin{equation}
\ket{w_{t;0}}_D=\sum_{j=2}^{N}\widetilde{P}_j\left(\omega_j^t\ket{w_{0;0}}-\eta\sum_{l=0}^{t-1}\omega_j^{t-l}\sum_{\beta=0}^{\tau-1}a_{\tau,\beta}\ket{\nabla F(w_{l;\beta})}\right).
\label{eq:transient_component_general}
\end{equation}
}

\noindent
The first component (Eq. (\ref{eq:consensus_component_general})) is the projection of the learning state onto the consensus subspace, where all node models are identical. It therefore represents a common trajectory shared by all nodes, rather than necessarily their actual model parameters. The second component (Eq. \ref{eq:transient_component_general})  captures deviations from this trajectory, arising from differences in the initial models and from heterogeneous local learning updates. While the contribution of initial disagreement decays over time, gradient-induced disagreement may persist because new local updates continuously introduce differences between nodes.

To make the role of network topology more explicit, we now focus on the case without momentum and with a single local step per communication round, corresponding to $\mu=0$ and $\tau=1$. The general expressions above remain valid for arbitrary $\tau$ and $\mu$.

\subsubsection{Consensus Trajectory}
\label{sec:consensus_trajectory}

For $\mu=0$ and $\tau=1$, the coefficients $a_{\tau,\beta}$ defined in Equation~\eqref{eq:momentum_coefficient} reduce to one, and the consensus component in Equation~\eqref{eq:consensus_component_general} simplifies to
\begin{equation}
\ket{w_{t;0}}_C=\widetilde{P}_1\left(\ket{w_{0;0}}-\eta\sum_{l=0}^{t-1}\ket{\nabla F(w_{l;0})}\right).
\label{eq:consensus_trajectory}
\end{equation}
Substituting $P_1=|1\rangle\langle\psi_1|$ and expanding the projection over individual nodes, we obtain the consensus component at node $i$:
\begin{equation}
\mathbf{w}_{t;0,C}^{(i)}=\sum_{k=1}^{N}\psi_1^{(k)}\left(\mathbf{w}_{0;0}^{(k)}-\eta\sum_{l=0}^{t-1}\nabla F_k(\mathbf{w}_{l;0}^{(k)})\right),
\label{eq:node_consensus_component}
\end{equation}
where $\psi_1^{(k)}$ is the $k$-th entry of the dominant left eigenvector. As expected, the expression is independent of the node index $i$. Although Equation~\eqref{eq:node_consensus_component} separates the consensus component from disagreement, their dynamics are not generally independent. Indeed, each local gradient is evaluated at the actual node model $\mathbf{w}_{l;0}^{(k)}=\mathbf{w}_{l;0,C}+\mathbf{w}_{l;0,D}^{(k)}$. Consequently, disagreement can indirectly affect the consensus trajectory through the local gradients, even though the non-dominant eigenmodes do not contribute directly to the consensus projection.

To determine the contribution of each node to the consensus trajectory, we need to characterize the dominant left eigenvector $\langle\psi_1|$ of $K$. As shown in the following theorem, the structure of $K=(D')^{-1}A'$ and the symmetry of $A'$ allow us to express its entries directly in terms of weighted node degrees, using the classical stationary-distribution characterization of reversible random walks~\cite{ref:chung_spectral}.

\begin{theorem}[Stationary distribution and consensus weights]
\label{thm:consensus_trajectory}
Under the assumptions of Lemma~\ref{lemma:K_properties}, let $d'_i=\sum_{j=1}^{N}A'_{ij}$ denote the weighted degree of node $i$, including its self-loop weight. The dominant left eigenvector, normalized such that $\braket{\psi_1}{1}=1$, satisfies
\begin{equation}
\psi_1^{(i)}=\frac{d'_i}{\sum_{k=1}^{N}d'_k}.
\label{eq:stationary_distribution}
\end{equation}
Consequently, the contribution of each node to the consensus trajectory is weighted proportionally to its weighted degree.
\end{theorem}

\begin{proof}
Since $K=(D')^{-1}A'$ and $A'$ is symmetric, we have $\mathbf{1}^T D'K=\mathbf{1}^T A'=\mathbf{1}^T D'$, where the last equality follows from $D'=\operatorname{diag}(A'\mathbf{1})$. Thus, $\mathbf{1}^T D'$ is a left eigenvector of $K$ associated with $\omega_1=1$. Since this eigenvalue is simple (Theorem~\ref{thm:dominant_eigenmode}), normalizing the eigenvector such that $\langle\psi_1|\mathbf{1}\rangle=1$ yields $\psi_1^{(i)}=d'_i/\sum_{k=1}^N d'_k$, proving the result.
\end{proof}

Theorem~\ref{thm:consensus_trajectory} establishes a direct relationship between network topology and the consensus trajectory. Unlike doubly stochastic averaging, which assigns equal influence to all nodes, DecAvg weights their contributions according to the stationary distribution of $K$. Under uniform neighborhood averaging, where all edge and self-loop weights equal one, Equation~\eqref{eq:stationary_distribution} reduces to
\begin{equation}
\psi_1^{(i)}=\frac{d_i+1}{2E+N}.
\label{eq:uniform_stationary_distribution}
\end{equation}
\textbf{Interpretation.} This result has an important implication: \emph{nodes with higher degrees exert greater influence on the consensus trajectory}. Consequently, the placement of data within the network matters, particularly when local data distributions are heterogeneous. Information held by highly connected nodes receives greater weight in the dominant component of the learning dynamics, whereas information held by peripheral nodes receives less weight.
For \emph{regular graphs} with uniform aggregation weights, all nodes have the same degree and therefore $\psi_1^{(i)}=1/N$. In this case, the consensus trajectory assigns equal influence to every node. 
Interestingly, approximately uniform stationary influence is not restricted to regular graphs. For \emph{Erdős--Rényi random graphs} $G(N,p)$, node degrees concentrate around their expected value as the average degree increases. In particular, when $Np\gg\log N$, the stationary influence weights satisfy $\psi_1^{(i)}=\frac{1}{N}\left(1+O\left(\sqrt{\frac{\log N}{Np}}\right)\right)$ uniformly across nodes with high probability (Theorem~\ref{thm:er_stationary_distribution}). Thus, although ER graphs are not exactly regular, their stationary influence distribution approaches uniformity in sufficiently dense regimes. This suggests that their consensus trajectories may resemble those of regular graphs when local gradient contributions are comparable.

Finally, in \emph{irregular graphs}, instead, the non-uniform stationary distribution introduces a topology-dependent weighting of local contributions. This distinction will be central to the analysis of non-IID data distributions in Section~\ref{sec:scenarios}.

\subsubsection{Disagreement Dynamics}
\label{sec:transient_corrections}

We now consider the contribution of the non-dominant eigenmodes. Under the same assumptions of $\mu=0$ and $\tau=1$, Equation~\eqref{eq:transient_component_general} becomes
\begin{equation}
\ket{w_{t;0}}_D=\sum_{j=2}^{N}\widetilde{P}_j\left(\omega_j^t\ket{w_{0;0}}-\eta\sum_{l=0}^{t-1}\omega_j^{t-l}\ket{\nabla F(w_{l;0})}\right).
\label{eq:transient_eq}
\end{equation}
This expression makes explicit the two sources of disagreement: differences between the initial models and heterogeneous local gradients. The contribution of initial disagreement decays geometrically, with a worst-case rate governed by $\rho=\max_{j\geq 2}|\omega_j|<1$. Gradient-induced disagreement, instead, may persist because local updates continuously introduce new differences between models. Its magnitude depends on how gradient heterogeneity projects onto the non-dominant eigenmodes and on the corresponding eigenvalues.

Importantly, gradient heterogeneity does not necessarily imply non-IID data. Differences between local gradients may also arise from stochastic gradient noise or differences between local model parameters, even when data are IID. As discussed in Section~\ref{sec:iid_data}, these differences can excite non-dominant eigenmodes and introduce topology dependence into the learning dynamics.

The spectral decomposition also allows us to quantify how network mixing limits the accumulation of disagreement. Let $\boldsymbol{\pi}$ denote the stationary distribution of $K$, and define the stationary-weighted norm of a stacked vector as $\|\mathbf{x}\|_\pi^2=\sum_{i=1}^{N}\pi_i\|\mathbf{x}^{(i)}\|^2$. Since $K$ is reversible with respect to $\boldsymbol{\pi}$, this norm provides a natural measure of disagreement for the generally non-symmetric mixing matrix. Stationary-weighted norms are also used in existing analyses of row-stochastic decentralized optimization~\cite{liu2026rowstochastic}. Here, we use this norm to characterize how spectral mixing and gradient disagreement jointly bound deviations from the consensus trajectory.
The following corollary establishes a bound on disagreement dynamics; its proof is provided in Appendix~\ref{app:proof_disagreement_bound}.

\begin{corollary}[Bound on disagreement dynamics]
\label{cor:disagreement_bound}
Under the assumptions of Lemma~\ref{lemma:K_properties}, consider DecAvg with $\mu=0$ and $\tau=1$. Let $\rho=\max_{j\geq2}|\omega_j|<1$, and define the gradient disagreement at round $l$ as
\begin{equation}
\ket{g_{l,D}}=(I_{Nd}-\widetilde{P}_1)\ket{\nabla F(w_{l;0})}.
\label{eq:gradient_disagreement}
\end{equation}
Equivalently, $\|\mathbf{g}_{l,D}\|_\pi^2=\sum_{i=1}^{N}\pi_i\|\nabla F_i(\mathbf{w}_{l;0}^{(i)})-\overline{\mathbf{g}}_{\pi,l}\|^2$, where $\overline{\mathbf{g}}_{\pi,l}=\sum_{k=1}^{N}\pi_k\nabla F_k(\mathbf{w}_{l;0}^{(k)})$ is the stationary-weighted mean gradient.
Then, the disagreement component of the stacked model parameters satisfies
\begin{equation}
\|\mathbf{w}_{t;0,D}\|_\pi\leq\rho^t\|\mathbf{w}_{0;0,D}\|_\pi+\eta\sum_{l=0}^{t-1}\rho^{t-l}\|\mathbf{g}_{l,D}\|_\pi.
\label{eq:disagreement_bound}
\end{equation}
If $\|\mathbf{g}_{l,D}\|_\pi\leq G_D$ for all $l$, then
\begin{equation}
\|\mathbf{w}_{t;0,D}\|_\pi\leq\rho^t\|\mathbf{w}_{0;0,D}\|_\pi+\eta G_D\frac{\rho(1-\rho^t)}{1-\rho}.
\label{eq:disagreement_bound_uniform}
\end{equation}
Consequently, $\limsup_{t\to\infty}\|\mathbf{w}_{t;0,D}\|_\pi\leq\eta G_D\rho/(1-\rho)$.
\end{corollary}
The bound concerns model disagreement immediately after averaging; subsequent local optimization may reintroduce disagreement, even when $\rho=0$, as in a complete graph. Corollary~\ref{cor:disagreement_bound} quantifies the interplay between network mixing and gradient heterogeneity. The factor $\rho$ characterizes the attenuation of disagreement, while $G_D$ bounds the gradient differences accumulated during training. Better mixing (smaller $\rho$) yields a tighter disagreement bound, whereas greater gradient heterogeneity increases it. Under homogeneous initialization, the initial disagreement term vanishes, leaving only the contribution of heterogeneous local gradients.

The bound is conditional on gradient disagreement remaining bounded. Since local gradients are evaluated at the current local models, $\|g_{l,D}\|_\pi$ also depends on disagreement accumulated in previous rounds. Thus, $G_D$ represents an assumption on gradient disagreement along the learning trajectory, rather than a quantity independently derived from the dynamics. This is related to the bounded-heterogeneity assumptions commonly adopted in decentralized SGD analysis~\cite{koloskova2020unified}, although it also captures differences arising from local model disagreement and stochastic gradient noise. Under IID data, gradient disagreement may remain small when stochastic gradient variance and model disagreement are controlled, whereas strongly heterogeneous data distributions may lead to larger values of $G_D$.

Moreover, the stationary-weighted norm assigns weight $\pi_i$ to node $i$, so deviations at nodes with low stationary influence contribute less to the global disagreement measure. Since the $i$-th block of $w_{t;0,D}$ represents the deviation from the stationary-weighted mean $\bar{w}_t=\sum_{k=1}^N\pi_k w^{(k)}_{t;0}$, we obtain the node-level bound
\begin{equation}
\|w^{(i)}_{t;0}-\bar{w}_t\|\leq\frac{\|w_{t;0,D}\|_\pi}{\sqrt{\pi_i}}.
\end{equation}
Under uniform neighborhood averaging, the factor $1/\sqrt{\pi_i}$ becomes $\sqrt{(2E+N)/(d_i+1)}$. Thus, the global bound provides weaker control over deviations at low-degree nodes, reflecting their smaller weight in the disagreement measure rather than necessarily larger deviations.

\section{Interplay Between Network Topology and Data Heterogeneity}
\label{sec:scenarios}

We now investigate how the interaction between network topology and data distribution affects decentralized learning. Assuming homogeneous model initialization, we first examine the conditions under which learning dynamics become independent of topology in IID settings. We then consider non-IID data distributions and derive a quantitative characterization of how the placement of class-specific information influences the consensus trajectory.

\subsection{Homogeneous Initialization}
\label{sec:homogeneous_initialization}

We assume that all nodes start from the same model parameters, i.e., $\mathbf{w}_{0;0}^{(i)}=\mathbf{w}_0$ for every node $i$. The initial stacked state can therefore be written as
\begin{equation}
\ket{w_{0;0}}=\ket{1}\otimes\ket{w_0}.
\label{eq:homogeneous_initialization}
\end{equation}
Since $\ket{1}$ spans the dominant eigenspace of $K$, homogeneous initialization implies $\widetilde{P}_1\ket{w_{0;0}}=\ket{w_{0;0}}$ and $\widetilde{P}_j\ket{w_{0;0}}=0$ for $j\geq 2$. Substituting these properties into Theorem~\ref{thm:eigenmode_decavg}, with $\mu=0$ and $\tau=1$, yields the following result.

\begin{theorem}[DecAvg dynamics under homogeneous initialization]
\label{thm:homogeneous_eigenvalue_problem}
Under the assumptions of Lemma~\ref{lemma:K_properties}, if all nodes initialize their models identically, the stacked model parameters satisfy
\begin{equation}
\begin{aligned}
\ket{w_{t;0}}={}&\ket{1}\otimes\ket{w_0}-\eta\sum_{l=0}^{t-1}\widetilde{P}_1\ket{\nabla F(w_{l;0})}\\
&-\eta\sum_{l=0}^{t-1}\sum_{j=2}^{N}\omega_j^{t-l}\widetilde{P}_j\ket{\nabla F(w_{l;0})}.
\end{aligned}
\label{eq:homogeneous_eigenvalue_problem_theorem}
\end{equation}
\end{theorem}

\begin{proof}
The result follows directly from Theorem~\ref{thm:eigenmode_decavg}, noting that homogeneous initialization eliminates the contribution of the initial state to every non-dominant eigenmode.
\end{proof}

\subsection{IID Data Distribution}
\label{sec:iid_data}

Under IID data, all nodes sample from the same underlying distribution and therefore share the same population objective. However, identical data distributions do not necessarily imply identical local gradients.

Consider first the idealized case in which all nodes compute identical gradients at every communication round, as would occur with exact gradient descent on the same population objective and homogeneous initialization. The stacked gradient can then be expressed as
\begin{equation}
\ket{\nabla F(w_{l;0})}=\ket{1}\otimes\ket{g_l},
\label{eq:iid_identical_gradients}
\end{equation}
where $\ket{g_l}$ denotes the common gradient at round $l$. Since this vector lies entirely in the consensus subspace, its projection onto every non-dominant eigenspace vanishes:
\begin{equation}
\widetilde{P}_j\ket{\nabla F(w_{l;0})}=0,\qquad j\geq 2.
\label{eq:iid_zero_disagreement}
\end{equation}
Substituting this property into Equation~\eqref{eq:homogeneous_eigenvalue_problem_theorem}, we obtain
\begin{equation}
\ket{w_{t;0}}=\ket{1}\otimes\left(\ket{w_0}-\eta\sum_{l=0}^{t-1}\ket{g_l}\right).
\label{eq:iid_consensus_dynamics}
\end{equation}
\textbf{Interpretation.} Equation~\eqref{eq:iid_consensus_dynamics} shows that all nodes follow the same gradient descent trajectory, independently of network topology. This follows because identical gradients excite only the dominant eigenmode, whose contribution is unaffected by the stationary influence weights.

In practice, IID data do not guarantee identical gradients. Finite local datasets and stochastic optimization introduce variability, while differences between local models can further increase gradient disagreement. These differences can excite non-dominant eigenmodes and introduce topology dependence through network mixing. Moreover, when gradients differ, the consensus trajectory itself may depend on topology because local contributions are weighted by the stationary distribution of $K$.

The magnitude of these effects depends on gradient variability. Consider two nodes evaluating stochastic gradients at identical model parameters, using independent minibatches of size $B$ drawn from the same distribution. If the expected squared deviation of individual-sample gradients from the population gradient is $\sigma^2$, then $\mathbb{E}[\|\mathbf{g}_i-\mathbf{g}_j\|^2]=2\sigma^2/B$~\cite{bottou2018optimization}. Thus, the root-mean-square gradient difference decreases as $1/\sqrt{B}$. Consequently, topology dependence under IID data is expected to diminish as gradient estimates become more accurate, provided that differences between local model parameters also remain small.

\subsection{Non-IID Data Distribution}
\label{sec:non_iid_data}

To characterize the interaction between network topology and heterogeneous data distributions, we consider a classification problem with $C$ classes. Let $\mathcal{D}_k^c$ denote the subset of node $k$'s dataset $\mathcal{D}_k$ belonging to class $c$, and let $p_c^{(k)}=|\mathcal{D}_k^c|/|\mathcal{D}_k|$ denote the corresponding class proportion. We define the class-conditional loss as $F_c^{(k)}(\mathbf{w})=\frac{1}{|\mathcal{D}_k^c|}\sum_{(x,y)\in \mathcal{D}_k^c}f(x,y;\mathbf{w})$, for classes with $|\mathcal{D}_k^c|>0$. Since each local objective is the average loss over its dataset, it can be written as $F_k(\mathbf{w})=\sum_{c=1}^{C}p_c^{(k)}F_c^{(k)}(\mathbf{w})$. Differentiating this expression, we obtain the following decomposition of the local gradient:
\begin{equation}
\nabla F_k(\mathbf{w}_{l;0}^{(k)})=\sum_{c=1}^{C}p_c^{(k)}\nabla F_c^{(k)}(\mathbf{w}_{l;0}^{(k)}).
\label{eq:local_class_gradient}
\end{equation}
Since each local objective is defined as an average over its dataset, the coefficients $p_c^{(k)}$ represent within-node class proportions rather than absolute sample counts. Under uniform neighborhood averaging, the stationary weights $\pi_k$ depend only on network topology. Consequently, the coefficient $\pi_k p_c^{(k)}$ assigned to each class-specific gradient is independent of the total local dataset size when class proportions are fixed. However, dataset cardinality can still affect the empirical class-conditional gradients and, consequently, their actual contribution to the consensus trajectory.

We define the stacked contribution of class $c$ at communication round $l$ as
\begin{equation}
\ket{g_{c,l}}=\sum_{k=1}^{N}p_c^{(k)}\ket{k}\otimes\ket{\nabla F_c^{(k)}(\mathbf{w}_{l;0}^{(k)})}.
\label{eq:class_gradient_contribution}
\end{equation}
The complete stacked gradient is therefore $\ket{\nabla F(w_{l;0})}=\sum_{c=1}^{C}\ket{g_{c,l}}$.
Substituting this decomposition into Equation~\eqref{eq:homogeneous_eigenvalue_problem_theorem} gives
\begin{equation}
\begin{aligned}
\ket{w_{t;0}}={}&\ket{1}\otimes\ket{w_0}-\eta\sum_{l=0}^{t-1}\sum_{c=1}^{C}\widetilde{P}_1\ket{g_{c,l}}\\
&-\eta\sum_{l=0}^{t-1}\sum_{j=2}^{N}\sum_{c=1}^{C}\omega_j^{t-l}\widetilde{P}_j\ket{g_{c,l}}.
\end{aligned}
\label{eq:non_iid_eigenvalue_problem_theorem}
\end{equation}

To characterize the consensus component, recall from Theorem~\ref{thm:consensus_trajectory} that the stationary influence weight of node $k$ is $\pi_k=d'_k/\sum_i d'_i$. The common consensus model $\mathbf{w}_{t;0,C} \in \mathbb{R}^{d}$ associated with the consensus trajectory therefore satisfies
\begin{equation}
\mathbf{w}_{t;0,C}=\mathbf{w}_0-\eta\sum_{l=0}^{t-1}\sum_{c=1}^{C}\sum_{k=1}^{N}\pi_k p_c^{(k)}\nabla F_c^{(k)}(\mathbf{w}_{l;0}^{(k)}).
\label{eq:non_iid_consensus_trajectory_theorem}
\end{equation}
\textbf{Interpretation.} Equation~\eqref{eq:non_iid_consensus_trajectory_theorem} reveals that the contribution of each class depends jointly on its distribution across nodes and on their stationary influence weights. In particular, the coefficient $\pi_k p_c^{(k)}$ captures the weight assigned to class $c$ through node $k$. Therefore, two systems with identical aggregate class proportions may exhibit different consensus trajectories if the classes are allocated to nodes with different degrees.

\subsubsection{Case Study: Localized Class Information}
\label{sec:localized_class}

We consider a binary classification problem with classes $c_1$ and $c_2$. Class $c_1$ is available only at a designated node $k^*$, while class $c_2$ is present at every node. This configuration represents an extreme form of data heterogeneity in which information about one class is localized within the communication network.
The local class proportions satisfy
\begin{equation}
p_{c_1}^{(k)}=\begin{cases}p_{c_1}^{(k^*)}, & k=k^*,\\0, & k\neq k^*,\end{cases}\qquad
p_{c_2}^{(k)}=1-p_{c_1}^{(k)}.
\label{eq:localized_class_distribution}
\end{equation}

To quantify the relative influence of the two classes on the consensus trajectory, we define the magnitude of their class-specific gradient contributions as
\begin{equation}
\Delta_c(l)=\left\|\sum_{k=1}^{N}\pi_k p_c^{(k)}\nabla F_c^{(k)}(\mathbf{w}_{l;0}^{(k)})\right\|.
\label{eq:class_contribution_magnitude}
\end{equation}
We then consider the ratio
\begin{equation}
R(l)=\frac{\Delta_{c_1}(l)}{\Delta_{c_2}(l)}.
\label{eq:class_contribution_ratio}
\end{equation}
The ratio $R(l)$ measures the relative magnitude of the two class-specific contributions to the consensus update, although it does not directly characterize their effects on learning accuracy or convergence time.

For the localized-class configuration, the numerator reduces to
\begin{equation}
\Delta_{c_1}(l)=\pi_{k^*}p_{c_1}^{(k^*)}\left\|\nabla F_{c_1}^{(k^*)}(\mathbf{w}_{l;0}^{(k^*)})\right\|.
\label{eq:localized_minority_contribution}
\end{equation}
To isolate the effect of topology-dependent weighting, we consider the case in which the class-$c_2$ gradients are approximately aligned and have comparable magnitudes across nodes. Under this assumption, the magnitude of their weighted sum can be approximated by the sum of their weighted magnitudes. Denoting their representative magnitude by $g_{c_2}(l)$, and the class-$c_1$ gradient magnitude at node $k^*$ by $g_{c_1}(l)$, we obtain
\begin{equation}
R(l)\approx\frac{\pi_{k^*}p_{c_1}^{(k^*)}}{1-\pi_{k^*}p_{c_1}^{(k^*)}}\frac{g_{c_1}(l)}{g_{c_2}(l)}.
\label{eq:non_iid_ratio}
\end{equation}

Defining $r(l)=g_{c_1}(l)/g_{c_2}(l)$ and substituting $\pi_{k^*}=d'_{k^*}/\sum_k d'_k$, the ratio becomes
\begin{equation}
R(l)\approx\frac{p_{c_1}^{(k^*)}}{\displaystyle\frac{\sum_{k=1}^{N}d'_k}{d'_{k^*}}-p_{c_1}^{(k^*)}}r(l).
\label{eq:non_iid_ratio_degrees}
\end{equation}
The approximation in Equations~\eqref{eq:non_iid_ratio}--\eqref{eq:non_iid_ratio_degrees} relies on the class-$c_2$ gradients being approximately aligned and having comparable magnitudes across nodes. Since these gradients are evaluated at the actual local models, disagreement induced by non-dominant eigenmodes can affect their alignment and magnitude, making the ratio indirectly dependent on spectral mixing. Finite-sample variability may introduce further directional differences and partial cancellation in the weighted gradient sum. Thus, the approximation is most informative when inter-node gradient differences are limited, while the exact class-contribution ratio remains valid without these assumptions.

Equation~\eqref{eq:non_iid_ratio_degrees} shows that the relative contribution of localized information depends on the stationary influence of node $k^*$, the local prevalence of class $c_1$, and the relative magnitudes of the class-specific gradients. Under the gradient-alignment and magnitude assumptions leading to Equation~\eqref{eq:non_iid_ratio_degrees}, placing class $c_1$ at a node with greater stationary influence increases its relative contribution to the consensus update, for fixed class proportions and relative gradient magnitudes. Thus, network topology can amplify or attenuate localized information depending on its placement.

\section{Experimental Validation}
\label{sec:topologies}

The experiments examine three implications of the analysis. First, we assess topology sensitivity under IID data. Second, using complete graphs to eliminate disagreement immediately after averaging, we examine the effects of network size and local class prevalence. Third, we place the localized class at a hub or a peripheral node of the same BA graph, keeping the mixing matrix fixed while changing data placement. The learning curves illustrate these mechanisms in controlled configurations. The time-rescaling comparisons are exploratory: the class-contribution ratio characterizes gradient contributions to the consensus update and does not establish a convergence-time scaling law.

\subsection{Experimental Setup}
\label{sec:experimental_setup}

Unless otherwise specified, all experiments use the same dataset, model architecture, optimization parameters, and evaluation methodology. The data allocation and number of participating nodes vary across scenarios and are specified in the corresponding subsections.

\medskip
\noindent\textbf{Dataset.} We use MNIST~\cite{lecun1998mnist} for handwritten digit classification. Its well-defined classes allow us to construct controlled IID and non-IID scenarios, varying network topology, size, and class allocation while keeping the learning task and model architecture fixed. Although MNIST is a relatively simple benchmark, this is appropriate for our objectives, as it allows us to isolate topology-dependent effects without introducing additional variability associated with more complex learning tasks. The data allocation across the different experiments is summarized in Table~\ref{tab:data_allocation}.

\begin{table}[t]
\centering
\caption{Data allocation in the experimental scenarios.}
\label{tab:data_allocation}
\footnotesize
\setlength{\tabcolsep}{4pt}
\begin{tabular}{lcc}
Data distribution  & Classes & \# images per node \\ \hline
IID & 0 to 9 & 10 \\\hline
Non-IID & 0 & 10 (all nodes)\\ &1 & 10 (only node $k^*$)\\ \hline
\end{tabular}
\end{table}

\medskip
\noindent\textbf{Model architecture.} Each node maintains a local multilayer perceptron (MLP) with three hidden layers of 512, 256, and 128 neurons, respectively, using ReLU activation functions. All nodes use the same architecture and initialize their model parameters identically, consistently with the assumptions of Section~\ref{sec:scenarios}.

\medskip
\noindent\textbf{Local optimization.} Each node performs full-batch gradient descent with learning rate $\eta=0.01$, one local step per communication round ($\tau=1$), and no momentum ($\mu=0$), consistently with the assumptions adopted in Sections~\ref{sec:consensus_disagreement}--\ref{sec:scenarios}.

\medskip
\noindent\textbf{Evaluation methodology.} We evaluate learning dynamics using the mean test accuracy and mean test loss across participating nodes. Both metrics are computed by averaging the corresponding values over all nodes, providing a consistent basis for comparing network topologies and data allocations.

The experiments are based on individual graphs and data realizations, as their purpose is to illustrate the mechanisms characterized by the spectral analysis rather than estimate average predictive performance across random configurations. For a fixed communication graph and sequence of local gradients, the spectral decomposition describes the learning dynamics exactly. The reported learning curves illustrate how the implications of this decomposition manifest in specific controlled configurations.

\medskip
\noindent\textbf{Communication topologies.} We consider five network topologies, divided into two families. Regular topologies include the complete graph, the ring, and random $4$-regular graphs. Heterogeneous topologies include Barabási--Albert (BA) graphs generated through preferential attachment with parameter $m=2$, and random graphs obtained by uniformly redistributing the edges of the corresponding BA networks among node pairs, without preserving individual node degrees. This produces a $G(N,|\mathcal{E}|)$ random-graph baseline with the same number of nodes and edges as the BA graph, allowing us to compare networks with identical density but different degree distributions. In particular, BA graphs exhibit heterogeneous degrees with highly connected hubs, whereas the rewired graphs provide a more homogeneous random-network baseline.

Throughout the experiments, we use uniform neighborhood averaging, with $\alpha_{ij}=1$ for every existing edge and self-loop. This choice isolates the effect of graph structure from that of explicitly assigned aggregation weights. The corresponding stationary influence weights are therefore $\pi_i=(d_i+1)/(2|E|+N)$, as derived in Equation~\eqref{eq:uniform_stationary_distribution}.
\subsection{IID Data: Limited Topology Dependence}
\label{sec:iid_experiments}

We first investigate the influence of network topology under IID data and homogeneous model initialization. 

\begin{figure}[!t]
    \centering
    \subfloat[]{\includegraphics[width=.46\linewidth,trim=0 0 0 13mm,clip]{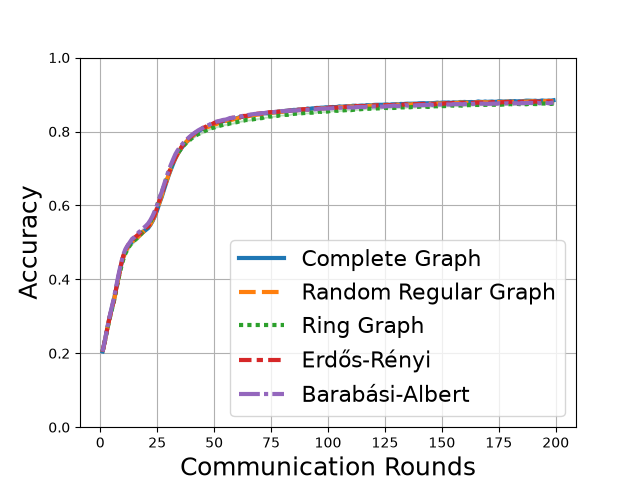}%
        \label{fig:iid_homo_acc}}\hfill
    \subfloat[]{\includegraphics[width=.46\linewidth,trim=0 0 0 13mm,clip]{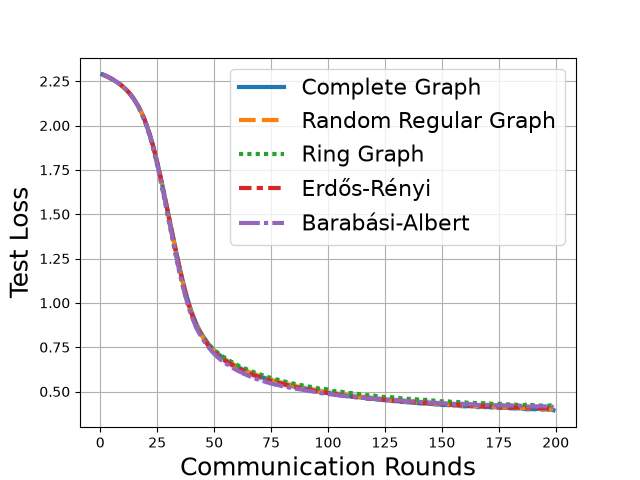}%
        \label{fig:iid_homo_loss}}
    \caption{Learning dynamics under IID data and homogeneous model initialization across five communication topologies: complete graph, ring, random regular graph, Erdős--Rényi graph, and Barabási--Albert graph.}
    \label{fig:all_topologies_iid}
\end{figure}

Figure~\ref{fig:all_topologies_iid} compares the evolution of test loss and accuracy across the five network topologies, each with 30 nodes. The learning curves are nearly indistinguishable, indicating that topology has little influence on the observed learning dynamics in this configuration. This result is consistent with the spectral interpretation developed in Section~6: when gradient disagreement is sufficiently small, neither the non-dominant eigenmodes nor the non-uniform stationary influence weights substantially alter the learning trajectory.

\setlength{\textfloatsep}{8pt plus 2pt minus 2pt}
\setlength{\floatsep}{6pt plus 2pt minus 2pt}
\setlength{\intextsep}{8pt plus 2pt minus 2pt}

\subsection{Non-IID Data: Topology-Dependent Class Influence}
\label{sec:non_iid_experiments}

We now investigate the localized-class scenario introduced in Section~\ref{sec:localized_class}. Class $c_1$ is available only at node $k^*$, while class $c_2$ is distributed across the network. This configuration allows us to examine how the stationary influence weight of the node holding the unique information affects the learning dynamics.

\subsubsection{Regular Topologies: Dilution of Localized Information}

We first consider regular graphs, for which the stationary distribution is uniform, $\pi_i=1/N$. Substituting this into Equation~\eqref{eq:non_iid_ratio}, the relative class-contribution ratio becomes
\begin{equation}
R(l)\approx\frac{p_{c_1}^{(k^*)}}{N-p_{c_1}^{(k^*)}}r(l).
\label{eq:non_iid_ratio_regular}
\end{equation}
For sufficiently large $N$, Equation~\eqref{eq:non_iid_ratio_regular} gives $R(l)\approx p_{c_1}^{(k^*)}r(l)/N$. Thus, for comparable class proportions and relative gradient magnitudes, the relative contribution of information localized at a single node decreases approximately as $1/N$. This dilution persists even when class $c_1$ dominates the local dataset at node $k^*$, showing that high local prevalence cannot compensate for the decreasing stationary influence as the network grows.

We investigate this effect using complete graphs, which provide a particularly useful reference case. With uniform neighborhood averaging, the mixing matrix of a complete graph satisfies $K=\mathbf{1}\mathbf{1}^T/N$, so all non-dominant eigenvalues are zero. Consequently, the disagreement introduced by local gradients is eliminated at each averaging step, allowing us to isolate the effect of the consensus trajectory.

\begin{figure}[!t]
    \centering
    \subfloat[Original learning curves.]{\includegraphics[width=.46\linewidth,trim=0 0 0 13mm,clip]{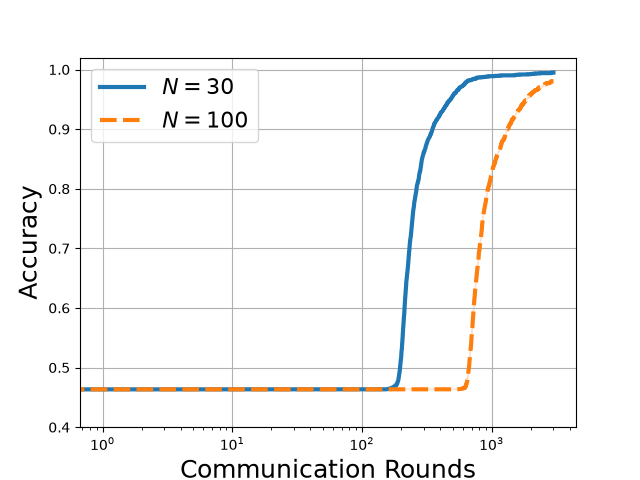}%
        \label{fig:complete_graph_different_sizes}}\hfill
    \subfloat[Time-rescaled learning curves.]{\includegraphics[width=.46\linewidth,trim=0 0 0 13mm,clip]{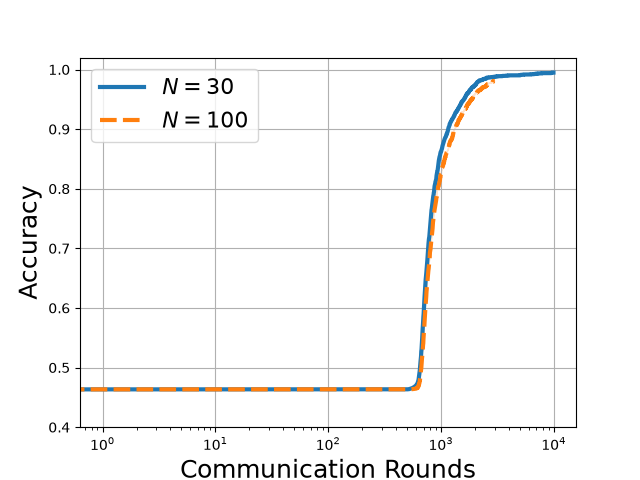}%
        \label{fig:complete_graph_different_sizes_shifted}}
    \caption{Effect of network size on minority-class learning in complete graphs with $N=30$ and $N=100$, under identical per-node data allocations.}
    \label{fig:complete_graph_size_comparison}
\end{figure}

Figure~\ref{fig:complete_graph_size_comparison} compares minority-class learning in complete graphs with $N=30$ and $N=100$, using identical per-node data allocations. The larger network learns the localized class more slowly, consistently with the stronger dilution predicted by Equation~\eqref{eq:non_iid_ratio_regular}. Assuming comparable class proportions and relative gradient magnitudes, the analytical ratio gives $R_{N=30}(l)/R_{N=100}(l)\approx 3.4$. We use this factor to rescale the time axis of the smaller network as an exploratory comparison. The resulting curves align closely, suggesting a relationship between the relative contribution of localized information and the time required to learn it. However, the class-contribution ratio does not establish a formal scaling law for convergence time.

\begin{figure}[!t]
    \centering
    \subfloat[]{\includegraphics[width=.46\linewidth,trim=0 0 0 13mm,clip]{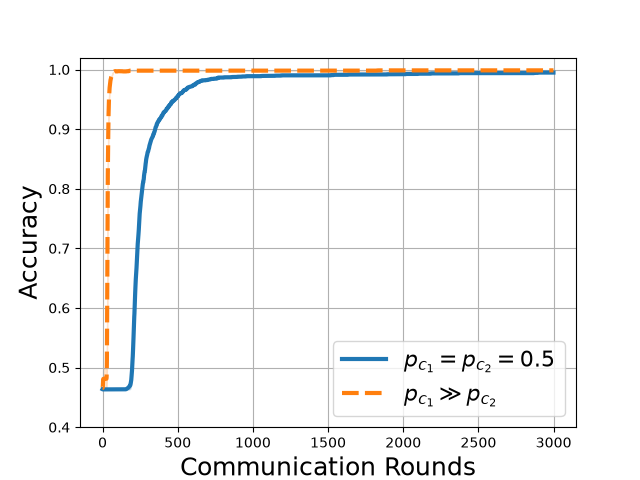}%
        \label{fig:complete_graph_different_data_distribution}}\hfill
    \subfloat[]{\includegraphics[width=.46\linewidth,trim=0 0 0 10mm,clip]{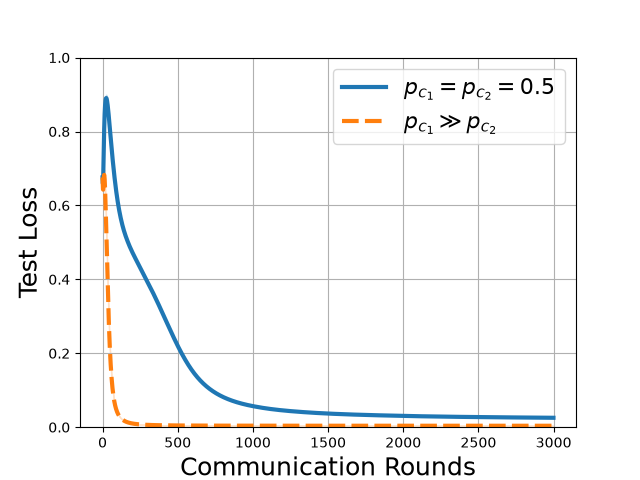}%
        \label{fig:complete_graph_different_data_loss}}
    \caption{Effect of local minority-class data availability on learning dynamics in a complete graph with $N=30$. The two configurations differ in the amount of class-$c_1$ data available at the node holding the localized class.}
    \label{fig:complete_graph_data_comparison}
\end{figure}

We next examine the effect of increasing the availability of class $c_1$ at node $k^*$ while keeping the network size fixed at $N=30$. Figure~\ref{fig:complete_graph_data_comparison} shows that increasing the number of minority-class samples accelerates learning and reduces the initial plateau. This behavior is consistent with the class-contribution ratio: increasing the number of minority-class samples raises their proportion within the local dataset, thereby increasing their relative weight in the consensus update, provided that relative gradient magnitudes remain comparable. The effect is mediated by local class proportions rather than absolute sample counts, although the latter may also affect the empirical gradients. The initial plateau may also reflect the evolving balance between class-specific gradients, although the ratio alone does not fully characterize the observed learning dynamics.

\subsubsection{Heterogeneous Topologies: Amplification and Suppression}

We now consider networks with heterogeneous degree distributions, focusing on Barabási--Albert (BA) graphs. Although both Erdős--Rényi (ER) and BA graphs exhibit degree heterogeneity, their stationary influence weights can differ substantially. As established in Theorem~\ref{thm:er_stationary_distribution} in the Appendix, sufficiently dense ER graphs have approximately uniform stationary distributions in the asymptotic regime $Np\gg\log N$. This result does not necessarily extend to small or sparse ER graphs, but it shows that degree heterogeneity need not imply strongly heterogeneous node influence. BA graphs, instead, typically exhibit more pronounced differences in node degrees, making them particularly suitable for investigating how the position of localized information within the network can amplify or suppress its contribution to the consensus trajectory.

To quantify this effect, we compare the stationary influence weight of node $k^*$ with the uniform weight $1/N$ assigned in a regular graph. We define the topology-induced amplification factor
\begin{equation}
A_{k^*}=N\pi_{k^*}=N\frac{d_{k^*}+1}{2|E|+N}.
\label{eq:topology_amplification}
\end{equation}
The factor $A_{k^*}$ measures the stationary influence of node $k^*$ relative to the uniform weight $1/N$: values above one indicate amplification, while values below one indicate suppression. Table~\ref{tab:stationary_influence} reports these factors for the graph realizations used in our experiments. For the BA realization considered here, the high- and low-degree nodes have degrees $d_{k^*}=18$ and $d_{k^*}=2$, respectively.

\begin{table}[t]
\centering
\caption{Stationary influence of the node holding the localized class in the experimental configurations.}
\label{tab:stationary_influence}
\begin{tabular}{lccc}
\hline
Configuration & $\pi_{k^*}$ & $A_{k^*}$ & $|E|$ \\
\hline
Complete graph & $1/N$ & $1$& 435 \\
BA (high-degree node) & 0.134 & 4.014 & 56 \\
BA (low-degree node) & 0.021 & 0.634 & 56 \\
\hline
\end{tabular}
\end{table}

We investigate the effect of information placement on a fixed BA graph realization, assigning the localized class either to a high-degree hub or to a low-degree node while preserving the data allocation. Since the communication matrix and its spectrum remain unchanged, this comparison isolates the effect of data placement from changes in network topology and spectral mixing rates. We also compare both configurations against a complete graph of the same size, which provides a reference with uniform stationary influence and perfect mixing.

\noindent\textbf{High-degree node.} When class $c_1$ is concentrated at a hub with $A_{k^*}>1$, its stationary influence weight is larger than in a regular graph of the same size. This increases the topology-dependent coefficient assigned to its gradient contribution, although the resulting effect on learning also depends on the gradient magnitudes and directions. According to Equation~\eqref{eq:non_iid_ratio}, increasing $\pi_{k^*}$ increases the relative contribution of class $c_1$, provided that the local class proportions and gradient magnitudes remain comparable.

\begin{figure}[t]
    \centering
    \subfloat[]{\includegraphics[width=0.29\linewidth,trim=0 0 0 13mm,clip]{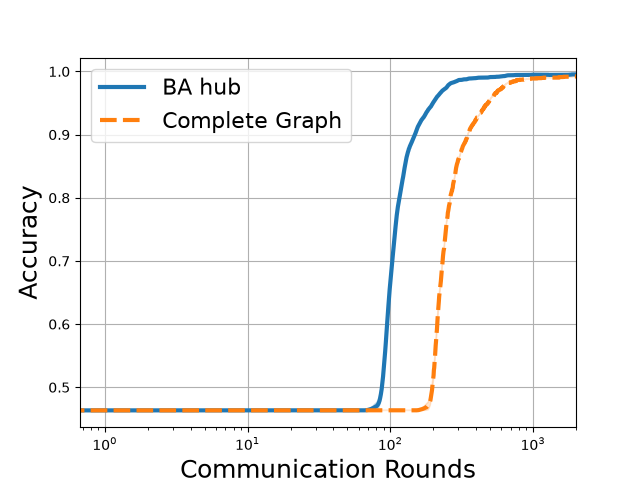}%
        \label{fig:ba_maxdeg_acc}}
    \hspace{0.02\linewidth}
    \subfloat[]{\includegraphics[width=0.29\linewidth,trim=0 0 0 13mm,clip]{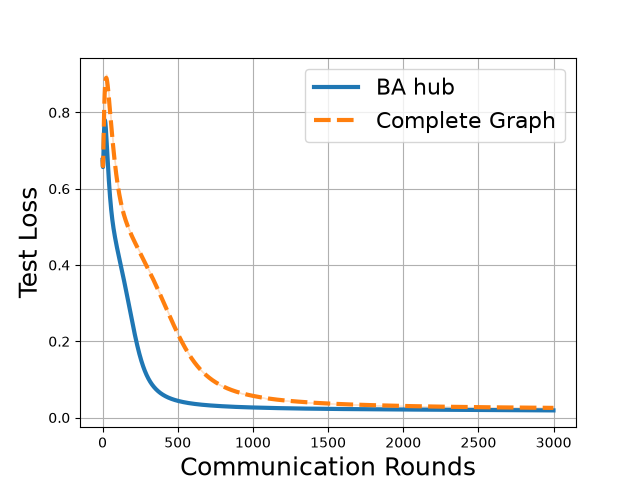}%
        \label{fig:ba_maxdeg_loss}}
    \hspace{0.02\linewidth}
    \subfloat[]{\includegraphics[width=0.29\linewidth,trim=0 0 0 13mm,clip]{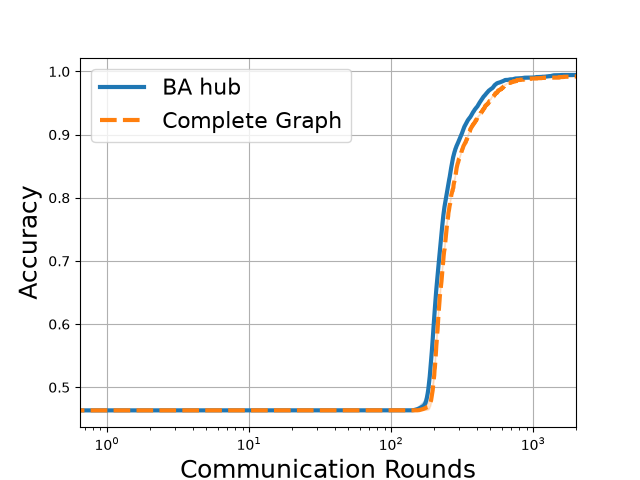}%
        \label{fig:ba_maxdeg_shifted_acc}}
    \caption{Learning dynamics for a Barabási--Albert graph with the minority class assigned to a high-degree node, compared with a complete graph under the same data allocation. Panels (a)--(b) show the original accuracy and loss curves, while panel (c) shows the accuracy curves after time rescaling.}
    \label{fig:ba_high_degree_special_node}
\end{figure}

Figure~\ref{fig:ba_high_degree_special_node} compares a BA graph with $N=30$ and preferential attachment parameter $m=2$ against a complete graph of the same size. The results show that the BA configuration learns the localized class more rapidly when the special node is a hub. The time-rescaled curves also exhibit a close alignment, suggesting that the amplification of the localized class contribution is associated with faster learning in this configuration.
This observation is consistent with the stationary influence analysis, although the class-contribution ratio alone does not establish a proportional relationship between influence amplification and convergence time. Since the two topologies also differ in their non-dominant spectra, the observed performance difference may reflect both stationary influence and disagreement dynamics.

\noindent\textbf{Low-degree node.} The opposite effect occurs when class $c_1$ is assigned to a peripheral node. In this case, its stationary influence weight may be smaller than $1/N$, reducing its contribution relative to a regular graph. The extent of this suppression is quantified by $A_{k^*}<1$.

\begin{figure}[!t]
    \centering
    \subfloat[Accuracy.]{\includegraphics[width=0.46\linewidth,trim=0 0 0 12mm,clip]{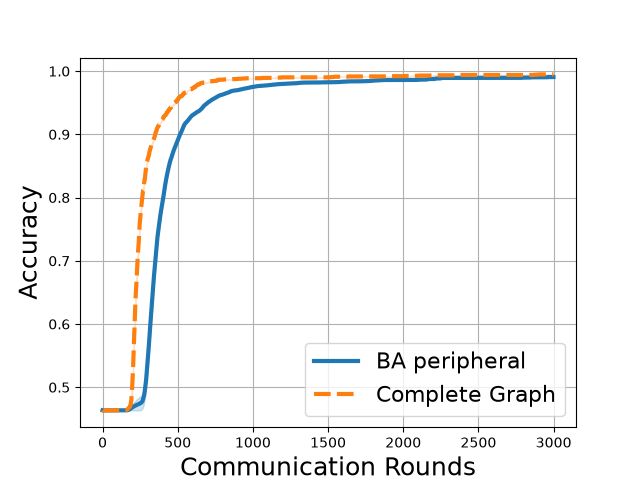}%
        \label{fig:ba_mindeg_acc}}\hfill
    \subfloat[Loss.]{\includegraphics[width=0.46\linewidth,trim=0 0 0 12mm,clip]{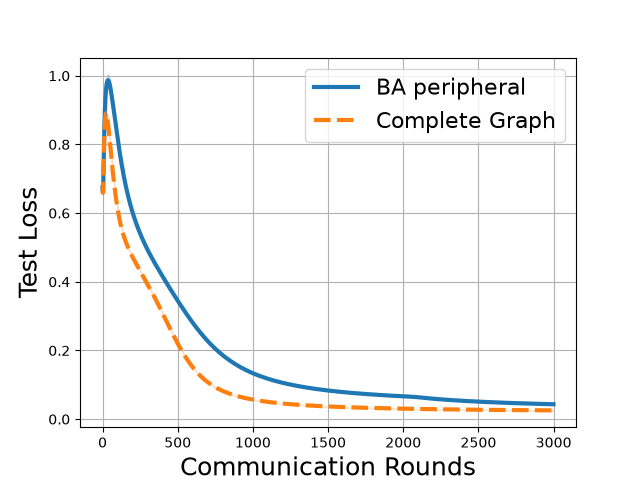}%
        \label{fig:ba_mindeg_loss}}
    \caption{Learning dynamics for a Barabási--Albert graph with the minority class assigned to a low-degree node, compared with a complete graph under the same data allocation.}
    \label{fig:ba_low_degree_special_node}
\end{figure}

Figure~\ref{fig:ba_low_degree_special_node} compares the learning dynamics of a BA graph with the localized class assigned to a minimum-degree node against those of a complete graph. The results illustrate how placing unique information at a peripheral node can delay its learning, despite using the same local data allocation.

Together, these experiments show that placing the same localized information at different nodes of an identical communication graph can produce substantially different learning dynamics. Since the mixing matrix and its spectrum remain unchanged, these differences cannot be attributed to changes in the network's spectral mixing rates. They are consistent with the topology-dependent stationary influence predicted by our analysis, although data placement also changes how local gradients excite the non-dominant eigenmodes.

\section{Conclusion}
\label{sec:conclusion}

We investigated how network topology and data heterogeneity jointly shape decentralized learning. Moving beyond convergence-oriented spectral analyses, we characterized how network structure determines the relative contribution of information held by different nodes. Our framework separates stationary node influence from disagreement dynamics and quantifies how the placement of localized information affects its contribution to the consensus trajectory. Experiments show limited topology dependence under IID data, whereas under non-IID data, learning dynamics are sensitive to network size and node connectivity.
These findings highlight that understanding decentralized learning requires considering not only how efficiently information propagates, but also how network structure weights the information being learned. This perspective is particularly relevant when data are unevenly distributed and communication topologies cannot be freely designed. Future work will extend the analysis to more general heterogeneous data distributions and to directed and time-varying networks.

\section*{Acknowledgments}

\begingroup
\catcode`\_=\active
\let_\textunderscore
This work was partially funded by the European Union within the National Programme 'Research, Innovation and Competitiveness' 2021-2027, projects SINTESI (Nr. MI_0000962) and AI4Nature (Nr. MI_0001001). C. Boldrini's and M. Conti's work was partly funded by the PNRR - M4C2 - Investimento 1.3, Partenariato Esteso PE00000013 - ``FAIR'', A. Passarella's and L. Valerio's work was partially supported by the European Union - Next Generation EU under the Italian National Recovery and Resilience Plan (NRRP), Mission 4, Component 2, Investment 1.3, CUP B53C22003970001, partnership on “Telecommunications of the Future” (PE00000001 - program “RESTART”). 

This research was supported by CHIST-ERA-19-XAI010 SAI projects, FWF (grant No. I 5205). JK acknowledges partial support by ERC grant No. 810115-DYNASNET.
\endgroup

The authors used OpenAI's ChatGPT and Anthropic's Claude to assist with language editing, grammar correction, and improving the clarity and readability of the manuscript. All scientific content was developed and validated by the authors.

\bibliographystyle{plain}
\bibliography{biblio}

@inproceedings{takezawa2026topology,
	author = {Takezawa, Yuki and Koloskova, Anastasia and Stich, Sebastian U},
	booktitle = {Proceedings of the 43rd International Conference on Machine Learning},
	editor = {Zhang, Tong and Dudik, Miroslav and Jaggi, Martin and Agarwal, Alekh and Li, Sharon and Schuurmans, Dale and Zhu, Jerry and Berkenkamp, Felix and Dong, Hanze and Bietti, Alberto},
	month = {06--11 Jul},
	pages = {118390--118420},
	publisher = {PMLR},
	series = {Proceedings of Machine Learning Research},
	title = {Improved Convergence Analysis of Topology Dependence in Decentralized {SGD}},
	url = {https://proceedings.mlr.press/v306/takezawa26a.html},
	volume = {306},
	year = {2026}}

@inproceedings{lebars2023topology,
	author = {Le Bars, Batiste and Bellet, Aur{\'e}lien and Tommasi, Marc and Lavoie, Erick and Kermarrec, Anne-Marie},
	booktitle = {Proceedings of the 26th International Conference on Artificial Intelligence and Statistics},
	pages = {1672--1702},
	publisher = {PMLR},
	series = {Proceedings of Machine Learning Research},
	title = {Refined Convergence and Topology Learning for Decentralized SGD with Heterogeneous Data},
	volume = {206},
	year = {2023}}

@inproceedings{liu2026rowstochastic,
	author = {Liu, Bing and Kong, Boao and Lu, Limin and Yuan, Kun and Zhao, Chengcheng},
	booktitle = {Proceedings of the 43rd International Conference on Machine Learning},
	pages = {77141--77187},
	publisher = {PMLR},
	series = {Proceedings of Machine Learning Research},
	title = {Row-Stochastic Matrices Can Provably Outperform Doubly Stochastic Matrices in Decentralized Learning},
	volume = {306},
	year = {2026}}

@inproceedings{liang2025rowstochastic,
	author = {Liang, Liyuan and Chen, Xinyi and Luo, Gan and Yuan, Kun},
	booktitle = {Proceedings of the 42nd International Conference on Machine Learning},
	pages = {37100--37130},
	publisher = {PMLR},
	series = {Proceedings of Machine Learning Research},
	title = {Achieving Linear Speedup and Near-Optimal Complexity for Decentralized Optimization over Row-stochastic Networks},
	volume = {267},
	year = {2025}}

@article{yuan2023heterogeneity,
	author = {Yuan, Kun and Alghunaim, Sulaiman A. and Huang, Xinmeng},
	journal = {Journal of Machine Learning Research},
	number = {280},
	pages = {1--53},
	title = {Removing Data Heterogeneity Influence Enhances Network Topology Dependence of Decentralized SGD},
	volume = {24},
	year = {2023}}

@inproceedings{huang2022sampletopology,
	author = {Huang, Yan and Sun, Ying and Zhu, Zehan and Yan, Changzhi and Xu, Jinming},
	booktitle = {Proceedings of the 39th International Conference on Machine Learning},
	pages = {9310--9345},
	publisher = {PMLR},
	series = {Proceedings of Machine Learning Research},
	title = {Tackling Data Heterogeneity: A New Unified Framework for Decentralized SGD with Sample-induced Topology},
	volume = {162},
	year = {2022}}

@article{palmieri2024topology,
	author = {Palmieri, Luigi and Boldrini, Chiara and Valerio, Lorenzo and Passarella, Andrea and Conti, Marco},
	doi = {10.1016/j.comnet.2024.110681},
	journal = {Computer Networks},
	pages = {110681},
	title = {Impact of Network Topology on the Performance of Decentralized Federated Learning},
	volume = {253},
	year = {2024}}

@article{badiemodiri2025initialisation,
	author = {Badie-Modiri, Arash and Boldrini, Chiara and Valerio, Lorenzo and Kert{\'e}sz, J{\'a}nos and Karsai, M{\'a}rton},
	doi = {10.1007/s41109-025-00737-4},
	journal = {Applied Network Science},
	pages = {53},
	title = {Initialisation and Network Effects in Decentralised Federated Learning},
	volume = {10},
	year = {2025}}

@misc{lecun1998mnist,
	author = {LeCun, Yann and Cortes, Corinna and Burges, Christopher J. C.},
	howpublished = {\url{https://yann.lecun.org/exdb/mnist/}},
	title = {The {MNIST} Database of Handwritten Digits},
	year = {1998}}

@article{bottou2018optimization,
	author = {Bottou, L{\'e}on and Curtis, Frank E and Nocedal, Jorge},
	journal = {SIAM review},
	number = {2},
	pages = {223--311},
	publisher = {SIAM},
	title = {Optimization methods for large-scale machine learning},
	volume = {60},
	year = {2018}}

@inproceedings{ref:mcmahan_fedavg,
	author = {McMahan, H. Brendan and Moore, Eider and Ramage, Daniel and Hampson, Seth and Arcas, Blaise Aguera y},
	booktitle = {Proceedings of the 20th International Conference on Artificial Intelligence and Statistics},
	pages = {1273--1282},
	series = {Proceedings of Machine Learning Research},
	title = {Communication-Efficient Learning of Deep Networks from Decentralized Data},
	volume = {54},
	year = {2017}}

@article{ref:nedic_olshevsky_consensus,
	author = {Nedi{\'c}, Angelia and Olshevsky, Alex},
	journal = {IEEE Transactions on Automatic Control},
	number = {3},
	pages = {601--615},
	title = {Distributed Optimization over Time-Varying Directed Graphs},
	volume = {60},
	year = {2015}}

@article{ref:boyd_fastmixing,
	author = {Boyd, Stephen and Diaconis, Persi and Xiao, Lin},
	journal = {SIAM Review},
	number = {4},
	pages = {667--689},
	title = {Fastest Mixing Markov Chain on a Graph},
	volume = {46},
	year = {2004}}

@book{ref:chung_spectral,
	author = {Chung, Fan R. K.},
	publisher = {American Mathematical Society},
	title = {Spectral Graph Theory},
	year = {1997}}

@article{sun_decentralized_2023,
	author = {T. Sun and D. Li and B. Wang},
	doi = {10.1109/TPAMI.2022.3196503},
	issn = {1939-3539},
	journal = {IEEE Trans. Pattern Anal. Mach. Intell.},
	number = {04},
	pages = {4289-4301},
	publisher = {IEEE Computer Society},
	title = {Decentralized Federated Averaging},
	volume = {45},
	year = {2023}}

@inproceedings{zhu2022topology,
	author = {Zhu, Tongtian and He, Fengxiang and Zhang, Lan and Niu, Zhengyang and Song, Mingli and Tao, Dacheng},
	booktitle = {International Conference on Machine Learning},
	organization = {PMLR},
	pages = {27479--27503},
	title = {Topology-aware generalization of decentralized sgd},
	year = {2022}}

@article{nedic2017achieving,
	author = {Nedic, Angelia and Olshevsky, Alex and Shi, Wei},
	journal = {SIAM Journal on Optimization},
	number = {4},
	pages = {2597--2633},
	publisher = {SIAM},
	title = {Achieving geometric convergence for distributed optimization over time-varying graphs},
	volume = {27},
	year = {2017}}

@inproceedings{lian_can_2017,
	author = {Lian, Xiangru and Zhang, Ce and Zhang, Huan and Hsieh, Cho-Jui and Zhang, Wei and Liu, Ji},
	booktitle = {Proceedings of the 34th International Conference on Machine Learning},
	pages = {2881--2890},
	series = {Proceedings of Machine Learning Research},
	title = {Can Decentralized Algorithms Outperform Centralized Algorithms? A Case Study for Decentralized Parallel Stochastic Gradient Descent},
	volume = {70},
	year = {2017}}

@inproceedings{koloskova2020unified,
	author = {Koloskova, Anastasia and Loizou, Nicolas and Boreiri, Sadra and Jaggi, Martin and Stich, Sebastian},
	booktitle = {International conference on machine learning},
	organization = {PMLR},
	pages = {5381--5393},
	title = {A unified theory of decentralized SGD with changing topology and local updates},
	year = {2020}}

@article{pu_asymptotic_2020,
	author = {Pu, Shi and Olshevsky, Alex and Paschalidis, Ioannis Ch.},
	copyright = {https://ieeexplore.ieee.org/Xplorehelp/downloads/license-information/IEEE.html},
	doi = {10.1109/MSP.2020.2975212},
	issn = {1053-5888, 1558-0792},
	journal = {IEEE Signal Processing Magazine},
	month = may,
	number = {3},
	pages = {114--122},
	shorttitle = {Asymptotic {Network} {Independence} in {Distributed} {Stochastic} {Optimization} for {Machine} {Learning}},
	title = {Asymptotic {Network} {Independence} in {Distributed} {Stochastic} {Optimization} for {Machine} {Learning}: {Examining} {Distributed} and {Centralized} {Stochastic} {Gradient} {Descent}},
	url = {https://ieeexplore.ieee.org/document/9084351/},
	urldate = {2025-12-10},
	volume = {37},
	year = {2020}}

@inproceedings{neglia2020decentralized,
	author = {Neglia, Giovanni and Xu, Chuan and Towsley, Don and Calbi, Gianmarco},
	booktitle = {International Conference on Artificial Intelligence and Statistics},
	organization = {PMLR},
	pages = {2348--2358},
	title = {Decentralized gradient methods: does topology matter?},
	year = {2020}}

@misc{dandi_data-heterogeneity-aware_2022,
	author = {Dandi, Yatin and Koloskova, Anastasia and Jaggi, Martin and Stich, Sebastian U.},
	doi = {10.48550/arXiv.2204.06477},
	month = apr,
	note = {arXiv:2204.06477 [cs]},
	publisher = {arXiv},
	title = {Data-heterogeneity-aware {Mixing} for {Decentralized} {Learning}},
	url = {http://arxiv.org/abs/2204.06477},
	urldate = {2025-12-10},
	year = {2022}}

@inproceedings{kempe2003gossip,
	author = {Kempe, David and Dobra, Alin and Gehrke, Johannes},
	booktitle = {44th Annual IEEE Symposium on Foundations of Computer Science, 2003. Proceedings.},
	organization = {IEEE},
	pages = {482--491},
	title = {Gossip-based computation of aggregate information},
	year = {2003}}

@inproceedings{assran2019stochastic,
	author = {Assran, Mahmoud and Loizou, Nicolas and Ballas, Nicolas and Rabbat, Mike},
	booktitle = {International Conference on Machine Learning},
	organization = {PMLR},
	pages = {344--353},
	title = {Stochastic gradient push for distributed deep learning},
	year = {2019}}

@misc{nedic_network_2018,
	author = {Nedi{\'c}, Angelia and Olshevsky, Alex and Rabbat, Michael G.},
	month = jan,
	note = {arXiv:1709.08765 [cs, math]},
	publisher = {arXiv},
	title = {Network {Topology} and {Communication}-{Computation} {Tradeoffs} in {Decentralized} {Optimization}},
	url = {http://arxiv.org/abs/1709.08765},
	urldate = {2023-09-11},
	year = {2018}}

\appendix

\renewcommand{\theequation}{\thesection.\arabic{equation}}
\setcounter{equation}{0}

\section{Proofs of the Theoretical Results}
\label{sec:proofs}

\subsection{Proof of Theorem~1: Matrix Formulation of DecAvg}

\begin{proof}
Recall that the communication step at round $t$ is described by the lifted interaction matrix $\widetilde{K}=K\otimes I_d$, where $K=(D')^{-1}A'$ is the node-level interaction matrix. The stacked model parameters immediately after averaging therefore satisfy
\begin{equation}
\mathbf{w}_{t;0}=\widetilde{K}\mathbf{w}_{t-1;\tau}.
\label{eq:appendix_averaging}
\end{equation}

We now express the result of the local optimization phase in terms of the model parameters at the beginning of the round. Under SGD with momentum, the local updates can be written in stacked form as
\begin{equation}
\mathbf{v}_{t;\beta+1}=\mu\mathbf{v}_{t;\beta}-\eta\nabla F(\mathbf{w}_{t;\beta}),\qquad \mathbf{w}_{t;\beta+1}=\mathbf{w}_{t;\beta}+\mathbf{v}_{t;\beta+1},
\label{eq:appendix_momentum}
\end{equation}
where $\mathbf{v}_{t;0}=0$ and $\beta=0,\ldots,\tau-1$.

Unrolling the momentum recursion gives
\begin{equation}
\mathbf{v}_{t;\beta+1}=-\eta\sum_{\alpha=0}^{\beta}\mu^{\beta-\alpha}\nabla F(\mathbf{w}_{t;\alpha}).
\end{equation}
Summing the parameter updates over the $\tau$ local steps, we obtain
\begin{equation}
\mathbf{w}_{t;\tau}=\mathbf{w}_{t;0}-\eta\sum_{\beta=0}^{\tau-1}\sum_{\alpha=0}^{\beta}\mu^{\beta-\alpha}\nabla F(\mathbf{w}_{t;\alpha}).
\end{equation}
Exchanging the order of summation yields
\begin{equation}
\mathbf{w}_{t;\tau}=\mathbf{w}_{t;0}-\eta\sum_{\alpha=0}^{\tau-1}\left(\sum_{k=0}^{\tau-1-\alpha}\mu^k\right)\nabla F(\mathbf{w}_{t;\alpha}).
\end{equation}

Since $\mu\in[0,1)$, the geometric sum satisfies
\begin{equation}
\sum_{k=0}^{\tau-1-\alpha}\mu^k=\frac{1-\mu^{\tau-\alpha}}{1-\mu}=a_{\tau,\alpha},
\end{equation}
where $a_{\tau,\alpha}$ is the momentum coefficient introduced in Equation~\eqref{eq:momentum_coefficient}. Therefore,
\begin{equation}
\mathbf{w}_{t;\tau}=\mathbf{w}_{t;0}-\eta\sum_{\alpha=0}^{\tau-1}a_{\tau,\alpha}\nabla F(\mathbf{w}_{t;\alpha}).
\label{eq:appendix_local_update}
\end{equation}

Substituting Equation~\eqref{eq:appendix_local_update} into Equation~\eqref{eq:appendix_averaging} gives
\begin{equation}
\mathbf{w}_{t;0}=\widetilde{K}\mathbf{w}_{t-1;0}-\eta\widetilde{K}\sum_{\alpha=0}^{\tau-1}a_{\tau,\alpha}\nabla F(\mathbf{w}_{t-1;\alpha}).
\end{equation}

Iterating this recursion over $t$ communication rounds, we obtain
\begin{equation}
\mathbf{w}_{t;0}=\widetilde{K}^{t}\mathbf{w}_{0;0}-\eta\sum_{l=0}^{t-1}\widetilde{K}^{t-l}\sum_{\alpha=0}^{\tau-1}a_{\tau,\alpha}\nabla F(\mathbf{w}_{l;\alpha}),
\end{equation}
which proves Theorem~1.
\end{proof}

\subsection{Proof of Lemma~\ref{lemma:K_properties}: Properties of the Interaction Matrix}

\begin{proof}
Recall that $K=(D')^{-1}A'$, where $A'$ is the symmetric weighted adjacency matrix including self-loops, and $D'=\operatorname{diag}(A'\mathbf{1})$ is the corresponding weighted degree matrix.

\medskip
\noindent\textsc{Part 1: Stochasticity and irreducibility.}

By construction, $K_{ij}=A'_{ij}/d'_i\geq0$. Moreover,
\begin{equation}
\sum_{j=1}^{N}K_{ij}=\frac{1}{d'_i}\sum_{j=1}^{N}A'_{ij}=1.
\end{equation}
Therefore, $K$ is row-stochastic.

Since the underlying graph is connected and all existing edges have positive weights, the directed graph associated with $K$ is strongly connected. Consequently, $K$ is irreducible. Furthermore, the positive self-loop weights imply $K_{ii}>0$ for every node, so the associated Markov chain is aperiodic.

Thus, $K$ is primitive and admits a unique stationary distribution.

\medskip
\noindent\textsc{Part 2: Relation to the graph Laplacian.}

Define the weighted graph Laplacian as $L_\alpha=D'-A'$. Since $A'=A_\alpha+\Gamma$ and $D'$ includes the corresponding weighted self-loop contributions, the diagonal self-loop terms cancel in $L_\alpha$.

Using $A'=D'-L_\alpha$, we obtain
\begin{equation}
K=(D')^{-1}(D'-L_\alpha)=I_N-(D')^{-1}L_\alpha.
\end{equation}
This establishes the relationship between the interaction matrix and the row-normalized weighted Laplacian.

\medskip
\noindent\textsc{Part 3: Diagonalizability and real spectrum.}

Consider the matrix
\begin{equation}
S=(D')^{-1/2}A'(D')^{-1/2}.
\end{equation}
Since $A'$ is symmetric and $D'$ is a positive diagonal matrix, $S$ is real and symmetric. Moreover,
\begin{equation}
K=(D')^{-1/2}S(D')^{1/2}.
\end{equation}
Therefore, $K$ is similar to the real symmetric matrix $S$.

By the spectral theorem, $S$ admits an orthonormal eigenbasis and has only real eigenvalues. Similarity preserves eigenvalues and diagonalizability, so $K$ is diagonalizable with a complete set of real eigenvalues.

This completes the proof of Lemma~\ref{lemma:K_properties}.
\end{proof}

\subsection{Proof of Theorem~\ref{thm:eigenmode_decavg}: Eigenmode Decomposition}

\begin{proof}
By Lemma~\ref{lemma:K_properties}, the interaction matrix $K$ is diagonalizable. Let $\{\omega_j\}_{j=1}^{N}$ denote its eigenvalues, and let $P_j=\ket{\phi_j}\bra{\psi_j}$ be the corresponding spectral projectors, with $\braket{\psi_i}{\phi_j}=\delta_{ij}$.

The spectral decomposition of $K$ gives
\begin{equation}
K^t=\sum_{j=1}^{N}\omega_j^tP_j.
\end{equation}

Since $\widetilde{K}=K\otimes I_d$, the corresponding decomposition in the stacked model space is
\begin{equation}
\widetilde{K}^{\,t}=\sum_{j=1}^{N}\omega_j^t\widetilde{P}_j,\qquad \widetilde{P}_j=P_j\otimes I_d.
\end{equation}

Substituting this expression into the matrix formulation of DecAvg in Theorem~1 yields
\begin{equation}
\begin{aligned}
\ket{w_{t;0}}={}&\sum_{j=1}^{N}\omega_j^t\widetilde{P}_j\ket{w_{0;0}}\\
&-\eta\sum_{l=0}^{t-1}\sum_{j=1}^{N}\omega_j^{t-l}\widetilde{P}_j\sum_{\beta=0}^{\tau-1}a_{\tau,\beta}\ket{\nabla F(w_{l;\beta})}.
\end{aligned}
\end{equation}
This is precisely the eigenmode decomposition stated in Theorem~\ref{thm:eigenmode_decavg}.
\end{proof}

\subsection{Proof of Corollary~\ref{cor:disagreement_bound}: Disagreement Bound}
\label{app:proof_disagreement_bound}

\begin{proof}
Let $\Pi=\operatorname{diag}(\boldsymbol{\pi})$, where $\boldsymbol{\pi}$ is the stationary distribution of $K$. Since $K=(D')^{-1}A'$ with symmetric $A'$, the detailed balance condition $\pi_iK_{ij}=\pi_jK_{ji}$ holds. Consequently, $\Pi K=K^T\Pi$, so $K$ is self-adjoint with respect to the weighted inner product $\langle\mathbf{x},\mathbf{y}\rangle_\pi=\mathbf{x}^T\Pi\mathbf{y}$.

The lifted operator $\widetilde{K}=K\otimes I_d$ is therefore self-adjoint with respect to the corresponding weighted inner product on $\mathbb{R}^{Nd}$, defined by $\Pi\otimes I_d$. Its spectral projectors are orthogonal in this inner product, and the operator norm of its restriction to the disagreement subspace is $\rho=\max_{j\geq2}|\omega_j|$.

Let $\widetilde{P}_D=I_{Nd}-\widetilde{P}_1$ denote the projector onto the disagreement subspace. For $\mu=0$ and $\tau=1$, Theorem~\ref{thm:eigenmode_decavg} gives
\begin{equation}
\mathbf{w}_{t;0,D}=\widetilde{K}^{t}\mathbf{w}_{0;0,D}-\eta\sum_{l=0}^{t-1}\widetilde{K}^{t-l}\mathbf{g}_{l,D},
\end{equation}
where $\mathbf{g}_{l,D}=\widetilde{P}_D\nabla F(\mathbf{w}_{l;0})$.

By the spectral properties of $\widetilde{K}$ on the disagreement subspace,
\begin{equation}
\|\widetilde{K}^{s}\mathbf{x}_D\|_\pi\leq\rho^s\|\mathbf{x}_D\|_\pi
\end{equation}
for every disagreement vector $\mathbf{x}_D$ and integer $s\geq0$.

Applying the triangle inequality yields
\begin{equation}
\|\mathbf{w}_{t;0,D}\|_\pi\leq\rho^t\|\mathbf{w}_{0;0,D}\|_\pi+\eta\sum_{l=0}^{t-1}\rho^{t-l}\|\mathbf{g}_{l,D}\|_\pi.
\end{equation}

If $\|\mathbf{g}_{l,D}\|_\pi\leq G_D$ for all $l$, the geometric sum satisfies
\begin{equation}
\sum_{l=0}^{t-1}\rho^{t-l}=\frac{\rho(1-\rho^t)}{1-\rho}.
\end{equation}
Substituting this expression gives Equation~\eqref{eq:disagreement_bound_uniform}. Taking the limit superior as $t\to\infty$ completes the proof.
\end{proof}

\subsection{Stationary Distribution of Erdős--Rényi Graphs}
\label{sec:er_stationary_distribution}

\renewcommand{\thetheorem}{\thesection.\arabic{theorem}}
\renewcommand{\theHtheorem}{\theHsection.\arabic{theorem}}
\setcounter{theorem}{0}

\begin{theorem}[Stationary distribution of Erdős--Rényi graphs]
\label{thm:er_stationary_distribution}
Let $G(N,p)$ be an Erd\H{o}s--R\'enyi random graph with $N$ nodes and edge probability $p=p(N)$. Under uniform neighborhood averaging with self-loops, suppose that $Np\gg\log N$ as $N\to\infty$, i.e., $\log N/(Np)\to 0$. Then, with high probability, the graph is connected and the stationary distribution of the interaction matrix $K$ satisfies
\begin{equation}
\psi_1^{(i)}=\frac{1}{N}\left(1+O\left(\sqrt{\frac{\log N}{Np}}\right)\right)
\label{eq:er_stationary_distribution}
\end{equation}
uniformly for all nodes $i$, with high probability.
\end{theorem}

\begin{proof}
In an Erdős--Rényi graph $G(N,p)$, each pair of distinct nodes is connected independently with probability $p$. Consequently, the degree $d_i$ of node $i$ follows a binomial distribution,
\begin{equation}
d_i\sim\mathrm{Bin}(N-1,p),
\end{equation}
with expected value $\mathbb{E}[d_i]=(N-1)p$ and variance $\mathrm{Var}(d_i)=(N-1)p(1-p)$.

From Theorem~\ref{thm:consensus_trajectory}, the stationary influence weight of node $i$ is
\begin{equation}
\psi_1^{(i)}=\frac{d_i+1}{\sum_{k=1}^{N}(d_k+1)}=\frac{d_i+1}{N(\bar d+1)},
\label{eq:er_stationary_exact}
\end{equation}
where $\bar d=\frac{1}{N}\sum_{k=1}^{N}d_k$ denotes the realized average degree. Rearranging, we obtain the exact identity
\begin{equation}
\psi_1^{(i)}=\frac{1}{N}\left(1+\frac{d_i-\bar d}{\bar d+1}\right).
\label{eq:er_stationary_deviation}
\end{equation}
Moreover, since $Np\gg\log N$ lies above the connectivity threshold of Erd\H{o}s--R\'enyi graphs, $G(N,p)$ is connected with high probability, justifying the application of Theorem~\ref{thm:er_stationary_distribution}.

Let $\bar d_{\mathrm{exp}}=(N-1)p$ denote the expected degree. Applying a Chernoff bound to each binomial degree $d_i$ and taking a union bound over all $N$ nodes, we obtain, with high probability,
\begin{equation}
\max_{1\leq i\leq N}|d_i-\bar d_{\mathrm{exp}}|=O\left(\sqrt{Np\log N}\right),
\label{eq:er_degree_concentration}
\end{equation}
provided that $Np\gg\log N$. Since $\bar d$ is the average of the realized degrees, the same bound implies
\begin{equation}
|\bar d-\bar d_{\mathrm{exp}}|\leq\max_i|d_i-\bar d_{\mathrm{exp}}|=O\left(\sqrt{Np\log N}\right).
\end{equation}

By the triangle inequality, it follows that
\begin{equation}
\max_i|d_i-\bar d|\leq\max_i|d_i-\bar d_{\mathrm{exp}}|+|\bar d-\bar d_{\mathrm{exp}}|=O\left(\sqrt{Np\log N}\right).
\end{equation}

Under the assumption $Np\gg\log N$, the relative degree fluctuations vanish, and $\bar d+1=\Theta(Np)$ with high probability. Therefore,
\begin{equation}
\max_i\left|\frac{d_i-\bar d}{\bar d+1}\right|=O\left(\sqrt{\frac{\log N}{Np}}\right).
\end{equation}

Substituting this bound into Equation~\eqref{eq:er_stationary_deviation} yields
\begin{equation}
\psi_1^{(i)}=\frac{1}{N}\left(1+O\left(\sqrt{\frac{\log N}{Np}}\right)\right)
\end{equation}
uniformly across all nodes, with high probability.
\end{proof}

\end{document}